\documentclass[10pt,twocolumn,letterpaper]{article}

\usepackage{iccv}      

\usepackage{fontawesome}
\usepackage{multirow}
\usepackage{arydshln}
\usepackage{comment}
\usepackage{subcaption}
\usepackage{booktabs}
\usepackage{pifont}
\definecolor{cvprblue}{rgb}{0.21,0.49,0.74}
\usepackage[pagebackref,breaklinks,colorlinks,allcolors=cvprblue]{hyperref}
\renewcommand{\paragraph}[1]{\noindent\textbf{#1}~}

\usepackage{xcolor}

\title{Scaling-up Perceptual Video Quality Assessment}

\author{Ziheng Jia\textsuperscript{1*}, Zicheng Zhang\textsuperscript{1*}, Zeyu Zhang$^1$, Yingji Liang$^4$, Xiaorong Zhu$^1$, Chunyi Li$^1$, Jinliang Han$^1$\\
Haoning Wu$^3$, Bin Wang$^2$, Haoran Zhang$^2$, Guanyu Zhu$^2$, Qiyong Zhao$^2$\\
Xiaohong Liu$^1$, Xiongkuo Min$^{1\diamondsuit}$, Guangtao Zhai$^1$\\
$^1$Shanghai Jiaotong University, $^2$Media Experience and Evaluation Lab, Huawei Techonologies\\$^3$Nanyang Technological University,$^4$East China Normal University
}
\begin{document}
\twocolumn[{%
\renewcommand\twocolumn[1][]{#1}%
\maketitle
\begin{center}
    \centering
    \vspace{-2.3em}
    \includegraphics[width=0.96\linewidth]{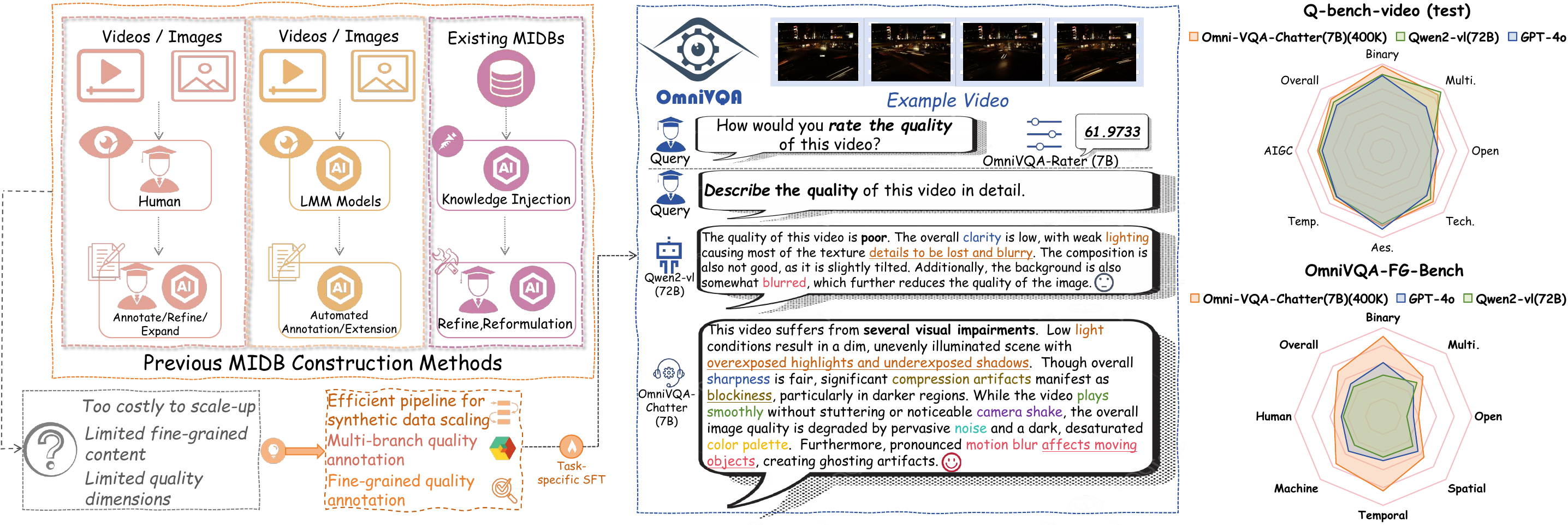}
    \vspace{-2pt}
    \captionof{figure}{\textbf{OmniVQA overview}. On the left side, it summarizes the existing MIDB construction paradigms in the visual quality assessment field. The main drawbacks of these methods lie in their excessive reliance on manual annotations or existing datasets, as well as the lack of comprehensive and diverse annotation dimensions. To address this, we have developed the OmniVQA datasets and models. On the right side, it shows examples of the OmniVQA model's application and the radar charts of its superior performance on two benchmarks.}
    \vspace{-2pt}
    \label{fig:intro}
\end{center}%
}]
\begin{abstract}
The data scaling law has been shown to significantly enhance the performance of large multi-modal models (LMMs) across various downstream tasks. However, in the domain of perceptual video quality assessment (VQA), the potential of scaling law remains unprecedented due to the scarcity of labeled resources and the insufficient scale of datasets. To address this, we propose 
\textbf{OmniVQA}, an efficient framework designed to efficiently build high-quality, human-in-the-loop VQA multi-modal instruction databases (MIDBs). We then scale up to create \textbf{OmniVQA-Chat-400K}, the largest MIDB in the VQA field concurrently. Our focus is on the technical and aesthetic quality dimensions, with abundant in-context instruction data to provide fine-grained VQA knowledge. Additionally, we have built the \textbf{OmniVQA-MOS-20K} dataset to enhance the model's quantitative quality rating capabilities.
We then introduce a \textbf{complementary} training strategy that effectively leverages the knowledge from datasets for quality understanding and quality rating tasks.  Furthermore, we propose the \textbf{OmniVQA-FG (fine-grain)-Benchmark} to evaluate the fine-grained performance of the models. Our results demonstrate that our models achieve state-of-the-art performance in both quality understanding and rating tasks.
\end{abstract}    
\vspace{-12pt}
\section{Introduction}

\label{sec:intro}

In the context of perceptual video quality assessment (VQA), the field currently focuses on two key tasks: quantitative quality rating and quality understanding. Quality rating refers to assigning a precise score to a video aligning with its human-labeled mean opinion score (MOS), while quality understanding involves providing detailed, qualitative feedback and analysis on the video's quality.
Recent advancements in large multi-modal models (LMMs) demonstrate the significant impact of data scaling laws on various downstream tasks \cite{wang2024llm,zhang2024scaling,zhou2023lima,zhang2024video}. 
However, in the field of perceptual video quality assessment (VQA), the benefits of data scaling have not been adequately investigated. It motivates a key hypothesis:
\textit{Scaling up VQA data can potentially improve model performance.}
Despite this potential, existing VQA datasets may struggle to fully leverage scaling due to limited labeling resources and insufficient data scale (shown in the supplementary materials (\textit{supp.})(Tab. \ref{tab:VQA_summary}).
Here, we would like to raise a fundamental query:

\textit{Why scaling up VQA data is a formidable challenge?}

Perceptual VQA is a task that \textbf{mimics human perception}, which inherently requires significant expert-level human involvement. As a consequence, the construction of large-scale datasets becomes not only resource-intensive but also time-consuming. Previous VQA multi-modal instruction databases (\textbf{MIDBs}) like \cite{wu2024q,huang2024aesexpert} 
utilize large language models (LLMs), such as \textit{GPT} \cite{radford2018improving}, to augment the human annotations. Nonetheless, the overall information gain remains marginal. Furthermore, this annotation methodology is deeply reliant on extensive human resources, thereby hindering its scalability.
To tackle these challenges, we introduce \textbf{OmniVQA}, a comprehensive framework by scaling perceptual VQA MIDB through a human-in-the-loop paradigm predominantly driven by machine annotation, culminating in the creation of the largest VQA MIDB to date, the \textbf{OmniVQA-Chat-400K}. This framework is built on $3$ branches that facilitate easy scalability while maintaining data quality. In addition, we propose \textbf{OmniVQA-MOS-20K}, a large-scale human-labeled VQA dataset specially for video quality rating tasks.


Considering the intrinsic interconnection between quality rating and understanding tasks, we further propose a task-specific \textbf{complementary} training strategy to effectively harness the quality knowledge embedded in datasets from both tasks, facilitating the training of LMMs. First, we train the model on one type of data (e.g., quality rating or understanding) and then finetune it using the remaining data. This approach enables the model to preserve and effectively integrate knowledge from both datasets.

Currently, there is no benchmark available for the fine-grained, in-context VQA quality understanding tasks. To thoroughly evaluate the in-context capabilities of our proposed model, we introduce the \textbf{OmniVQA-FG (fine-grain) Benchmark}, which is meticulously designed to assess the model's performance in spatiotemporal, local fine-grained video quality understanding and description.

Our contributions are threefold:
\begin{itemize}
    \item  We propose an effective and comprehensive data collection pipeline that supports the creation of large-scale, high-quality VQA MIDBs, and we then develop the OmniVQA-Chat-400K, with a focus on $3$ key branches: technical quality, aesthetic quality, and in-context analysis. We also propose OmniVQA-MOS-20K, a large-scale human-labeled UGC video subjective scoring dataset.
    \item The proposed OmniVQA models achieve state-of-the-art (SOTA) performance by training LMMs with the task-specific complementary training strategy for both quality rating and understanding (general and fine-grained) tasks.
    \item We introduce the OmniVQA-FG-Benchmark, a machine-human-mixed-annotated benchmark designed to evaluate fine-grained spatiotemporal quality understanding performance in both synthetic and real-world scenarios.
\end{itemize}

\vspace{-3pt}
\section{Related Works}
\label{sec:related}

\subsection{Perceptual video quality assessment}
Perceptual VQA initially focuses on quantitative video quality rating \cite{min2024perceptual}, primarily aiming to fit the human-labeled MOSs in public datasets such as \cite{ghadiyaram2017capture,nuutinen2016cvd2014,sinno2018large,wang2019youtube,li11662comparative,ying2021patch,duanmu2018quality,duanmu2016quality,bampis2021towards}. Existing works on this task includes handcrafted-feature-based methods \cite{mittal2012making,mittal2012no,tu2021ugc,duanmu2023bayesian,korhonen2019two,tu2021rapique,tu2021ugc,bampis2018feature}, deep neural network (DNN)-based approaches \cite{li2022blindly,li2019quality,liu2018end,wang2021rich,sun2022deep,sun2024analysis,wen2024modular,wu2023neighbourhood,wu2023discovqa,wu2023exploring,wu2023towards}, and LMM-based models \cite{wu2024q1,ge2024lmm,jia2024vqa}. Recently, video quality understanding has become a new emerging research field. \textbf{Q-bench-video} \cite{zhang2024q} is the first comprehensive benchmark for evaluating the capabilities of LMMs in general video quality understanding tasks. The \textbf{VQA\textsuperscript{2}-Assistant} \cite{jia2024vqa} is the first proprietary LMM capable of video quality understanding and chatting. However, existing works on perceptual VQA are almost entirely confined to the overall quality assessment of videos. There is almost no work addressing fine-grained tasks such as spatiotemporal local distortions retrieval and description, thereby leaving significant room for further research.
\begin{figure*}[h]
  \centering
  \includegraphics[width=0.98\linewidth]{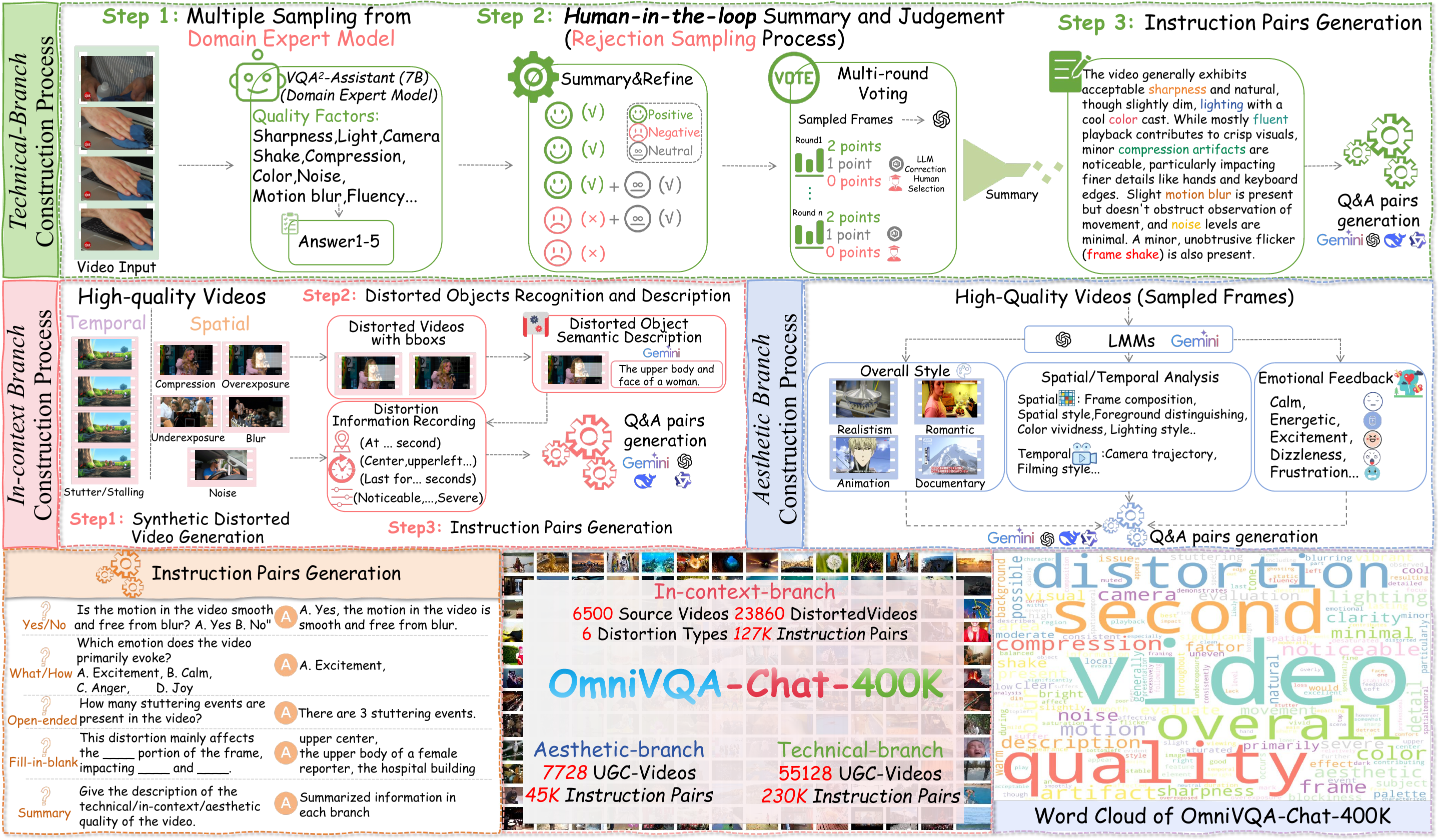}
    \vspace{-8pt}
   \caption{Data construction pipeline of \textbf{OmniVQA-Chat-400K}.}
   \vspace{-10pt}
   \label{fig:pipeline}
\end{figure*}
\subsection{MIDBs for visual quality assessment}
\label{MIDB}
In the field of LMMs for visual quality assessment, recent works have proposed various approaches to constructing MIDBs. These approaches can be broadly divided (shown in the part of Fig. \ref{fig:intro}):

\begin{enumerate}
    \item \textbf{Human annotators as perceivers }\cite{wu2024q,jia2024vqa,you2024depicting,huang2024aesexpert,zhou2024uniaa}: Human annotators directly serve as perceivers of videos/images, and the data is obtained directly through manual annotations and refinement/rewrite by LLMs. The major drawback is the significant cost of human labor and time, which significantly hampers the efficient scalability of MIDBs.
    \item \textbf{General LMMs as perceiver and annotator (distillation-like approach)}\cite{you2024descriptive}: General LMMs(e.g., \textit{GPT-4o} \cite{achiam2023gpt}, \textit{Gemini-1.5} \cite{team2024gemini}) directly serve as perceivers and annotators of image/video. Its primary limitation lies in the sub-optimal data quality, constrained by distillation from teacher models that are not proprietary within this domain.
    \item \textbf{Knowledge injection }\cite{wu2024q1,chen2024grounding,wu2024towards,chen2024q}: This approach involves injecting task-specific knowledge into existing MIDBs to adapt them for new downstream tasks with subsequent LLM refinement and reformulation. The deficiency with this method is that the scale of the new MIDB is heavily contingent upon the original MIDB, thus limiting the efficient expansion of the data scale.
\end{enumerate}
In addition to the aforementioned drawbacks, these previous MIDBs also share the following common issues. First, these datasets primarily focus on the annotation of the technical \cite{wu2024q,jia2024vqa,you2024depicting,you2024descriptive,wu2024q1,wu2024towards} or aesthetic quality \cite{huang2024aesexpert,zhou2024uniaa} of images/videos while lacking comprehensive annotations integrating multiple quality dimensions and factors. Moreover, the majority of these MIDBs \cite{wu2024q,jia2024vqa,you2024depicting,huang2024aesexpert,zhou2024uniaa,you2024descriptive,wu2024q1,wu2024towards} emphasize an overall description or understanding of image/video quality but lack spatiotemporal fine-grained annotation and evaluation. These issues have just become the inspiration for our work.

\vspace{-10pt}
\section{The OmniVQA-Chat-400K}
\label{OmniVQA-Chat-400K}
\vspace{-3pt}

\begin{figure*}[t]
  \centering
  \includegraphics[width=0.98\linewidth]{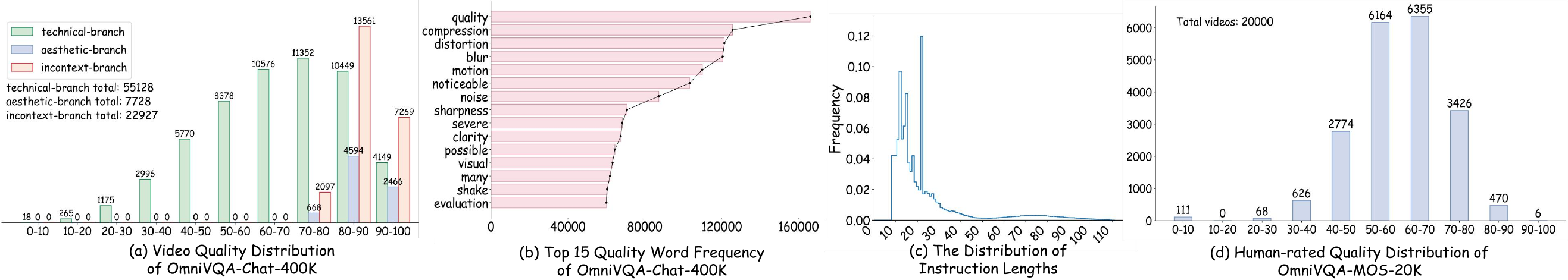}
    \vspace{-2pt}
   \caption{Statistical information of  \textbf{OmniVQA-Chat-400K} and  \textbf{OmniVQA-MOS-20K} .}
   \vspace{-15pt}
   \label{fig:statistical}
\end{figure*}

 Considering the deficiencies in the existing MIDBs, we propose that an effective perceptual VQA MIDB construction pipeline should encompass: \textbf{(1) Collaboration annotation using multiple models with human-in-the-loop. (2) Spanning multiple quality dimensions of video, along with spatiotemporal fine-grained data.}
Driven by these motivations, we construct the \textbf{OmniVQA-Chat-400K}, currently the largest and most comprehensive perceptual VQA MIDB. The construction pipeline is illustrated in Fig. \ref{fig:pipeline} and detailed in \textit{supp.} (Sec. \ref{process}), and some statistical information is shown in Fig. \ref{fig:statistical}.
\vspace{-5pt}
\subsection{Candidate video pool construction}
\vspace{-5pt}
We select $100,000$ videos from a large-scale online in-the-wild user-generated content (UGC) video dataset \cite{chen2024panda} to serve as the candidate video pool for subsequent selection. We impose the constraint that the length of all candidate videos must be in the range of \textit{[3,15)} seconds. We then utilize $4$ state-of-the-art objective video quality rating methods \cite{wu2023neighbourhood,wu2023exploring,wu2024q1,sun2024analysis} to label the objective quality labels for the candidate videos. To ensure consistency in the scale of the scores from each scoring method, they are first converted to $[0,100)$. Subsequently, the scores are averaged for each video as the objective quality label.
\subsection{The technical branch}
The \textbf{technical branch} is the primary component of the OmniVQA-Chat-400K. The videos in this branch are randomly selected from the candidate video pool to ensure the diversity of video content and quality. 

Each video is annotated across $8$ quality factors: \textbf{sharpness, light, compression, color, noise, fluency, motion blur}, and \textbf{camera shake}. We employ an aggregation strategy, where annotations for each quality factor are initially performed individually and subsequently combined (shown in the \textbf{upper} part of Fig.\ref{fig:pipeline}).

We propose a novel annotation paradigm based on \textbf{rejection sampling (RS)} for the annotation of each quality factor. The process begins by performing multiple coarse annotations using expert VQA LMM. This process forms the \textbf{suggested distribution}. Subsequently, unlike the common practice in reinforcement learning (RL) for RS \cite{liustatistical,khaki2024rs}, we do not only select the sampling result with the highest reward  determined by the reward model. Instead, we employ a heuristic method to fully utilize the effective information in all the sampled results. This method filters and summarizes the $N$ samples obtained through the first sampling with reasoning LLMs, and a voting mechanism using general LMMs is used to decide the next process. In this case, the general LMMs act as a \textbf{judger} rather than an annotator. We argue that substituting the challenging direct annotation task with the relatively simpler voting-based decision for general LMMs offers a more practical distillation approach. This method is more likely to guide the general LMM toward generating more accurate and definitive responses.

\paragraph{Multiple coarse annotations.}  We use \textit{VQA\textsuperscript{2}-Assistant (7B)} \cite{jia2024vqa} as the expert model for sampling.  We set \(N = 5\), meaning that for each quality factor, we pose $5$ questions with the same fundamental meaning but varying sentence structures and record the model's responses.

\paragraph{LLM summary.} We use a reasoning model to conduct the LLM summary process. The information in the responses obtained before is categorized into $3$ types:
\begin{enumerate}
    \item \textbf{Positive answers} are those that have similar meanings in the responses and appear in at least $3$ of the $5$ responses. These answers are then merged into the summary.
    \item \textbf{Negative answers}, which are contradicted in meaning to the positive answers, are excluded from the summary.
    \item If any responses contain additional \textbf{neutral} information, it should be included in the summary.
\end{enumerate}
For example, for the question \textit{``How is the sharpness of this video?"}, if the $5$ responses are: \textit{``Poor"} (positive),  \textit{``The sharpness is relatively poor"} (positive), \textit{``Poor, with degraded facial details"} (positive with neutral), \textit{``Good, however, the facial details are slightly lost"} (negative with neutral), \textit{``Excellent"} (negative).
The final summarized response would be:  
\textit{\textbf{``The sharpness is poor with degraded human facial details."}}

\paragraph{Voting and post-processing.}
We use \textit{sota LMM} to perform voting to decide the post-processing method. We input the keyframe sequence sampled from the annotated video (1fps) along with the prior summary of each quality factor. We then prompt it to conduct $3$ voting rounds for each quality factor, assessing the accuracy and relevance of the provided summaries and assigning a score in $(2,1,0)$. Then, the post-processing method is determined and implemented on the given quality factor summaries. If any round of voting scores is $0$, human experts are required to intervene and make a decision between the quality summary and it's modified summary for that round.
The detailed scoring criteria and the corresponding post-processing methods, including the human-in-the-loop selection process, are detailed in the \textit{supp.} (Fig. \ref{fig:branch-material-p2} and Sec. \ref{Human-in-the-loop}).



\paragraph{Instruction pairs generation.}
We use \textit{Openai-o1} to merge the annotations of all quality factors, resulting in a video-level quality summary. Based on this summary, we ask the model to generate $3$ question-answer (Q\&A) pairs related to specific quality factors. The questions may take the following forms:  binary choice questions, multiple-choice (single answer) questions, and Open-ended questions.

To fully leverage the overall quality summary, we add an extra question for each video:
\textit{``Please describe the overall quality of this video, please evaluate as many quality factors as possible."}
The video-level overall quality summary is then provided as the answer to this question.
\vspace{-3pt}
\subsection{The in-context branch}
\label{in-context}
The \textbf{in-context branch} is designed to augment the model's ability to identify fine-grained local spatiotemporal quality issues in videos. To minimize the influence of inherent distortions, we choose $6,500$ source videos from the candidate pool with objective quality labels above $70$. The annotation process is presented in the \textbf{middle left} part in Fig. \ref{fig:pipeline}.

\paragraph{Synthetic distorted video generation.}
We manually add local spatiotemporal distortions to the selected videos. Spatial distortions include \textbf{overexposure}, \textbf{underexposure}, \textbf{blur}, \textbf{compression artifacts}, and \textbf{noise}, while temporal distortions refer to \textbf{video stuttering}. 

Spatial distortions are randomly added to a rectangular region within the frame, covering $1/4$ of the frame area. 
The duration of spatial distortion is randomly assigned an integer value between $1$ and $3$ seconds, with the starting time of the synthetic distortion also determined as an integer. The intensity of the distortion is categorized into $3$ levels: \textbf{noticeable}, \textbf{relatively severe}, and \textbf{severe}. The $5$ distortion types are added sequentially for each source video, with the number of distortions fixed at $1$. Finally, we record the spatial distortion's start time, duration, distortion type, intensity, and location.  For the video stutter distortion, we remove the entire second following a randomly selected integer second and then duplicate the frames from that for an additional second, thereby creating a \textbf{frame freeze} effect. Each source video ultimately has $1$ to $3$ randomly inserted instances of video stutter events. More detailed video generation processes are provided in the \textit{supp.} (Sec. \ref{Distorted})). We  manually filter out the generated videos in which the synthetic spatiotemporal distortion is not visibly perceptible.



\paragraph{Distorted objects recognition and description.}
To enhance the LMM‘s ability to capture and describe the semantics of the local distortion regions, we add a highlighted bounding box (bbox) around the distorted rectangle area in the generated videos. The keyframes of the distorted videos with the bbox, are then input into the \textbf{sota LMM} model, which is tasked with describing the main semantic objects within the bbox with the following criteria: a semantic object is deemed valuable for annotation only if \textit{it is fully contained within the bbox and occupies the primary region, exhibits a distinct contrast from the surrounding areas, and remains within the bbox throughout its presence within the distorted period}.


\paragraph{Instruction pairs generation.}
We then generate $5$ instruction Q\&A pairs for each spatial and temporal distorted video based on the summarized video distortion information. The Q\&A pairs must focus entirely on the spatiotemporal local distortions of the video. If the video information contains the annotated distorted semantic object, at least one of the generated Q\&A pairs must be related to this. Here, we introduce an additional form of Q\&A pair, which is the cloze completion for the local distortion information.

Similar to the technical branch, each video also includes a summarizing question, which is:
\textit{``Please describe the information of the spatiotemporal local distortions of the video."}
The answer to this question is set to be the summarized video’s local distortion information.
\begin{figure}[t]
  \centering
  \includegraphics[width=\linewidth]{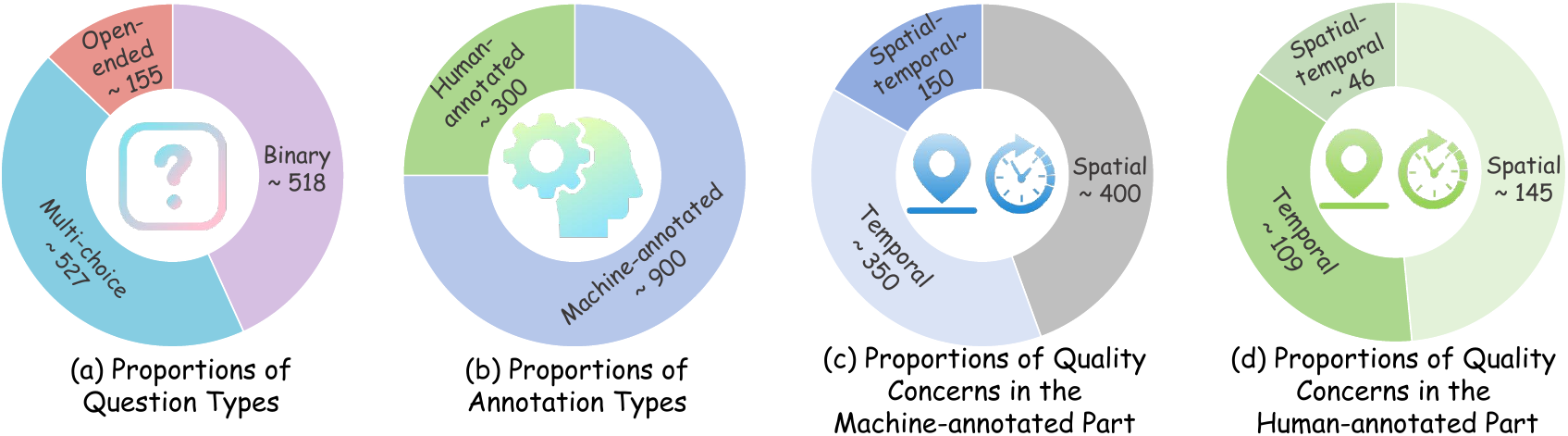}
    \vspace{-12pt}
   \caption{Statistical information of the OmniVQA-FG-Bench.}
   \vspace{-11pt}
   \label{fig:benchstatistic}
\end{figure}
\vspace{-3pt}
\subsection{The aesthetic branch}
The aesthetic branch aims to enhance the diversity and comprehensiveness of the MIDB.
To prevent severe technical distortions from affecting the extraction of aesthetic features, we also select videos with objective quality scores above $70$. Then, we directly input keyframes sampled from the videos into \textit{sota LMM} to annotate the aesthetic quality.

Inspired by \cite{huang2024aesexpert}, the annotation is conducted from $3$ aspects (illustrated in the \textbf{middle right} of Fig. \ref{fig:pipeline}):
\begin{enumerate}
    \item \textbf{Aesthetic Style}: Label the aesthetic style of the video.
    \item \textbf{Spatiotemporal Analysis}: Perform a detailed aesthetic analysis from the spatial and temporal perspectives.
    \item \textbf{Emotional Repercussions}: Provide the emotional experience that the video may evoke in viewers.
\end{enumerate}
The machine annotation for each video is then summarized, and $6$ instruction Q\&A pairs are derived from this. 

Each video also includes a summarizing Q\&A pair with the corresponding question \textit{``Please give a detailed description of the aesthetic effects of the video."} with all the annotated aesthetic information as the corresponding answer.

\begin{figure}[t]
  \centering
  \includegraphics[width=0.98\linewidth]{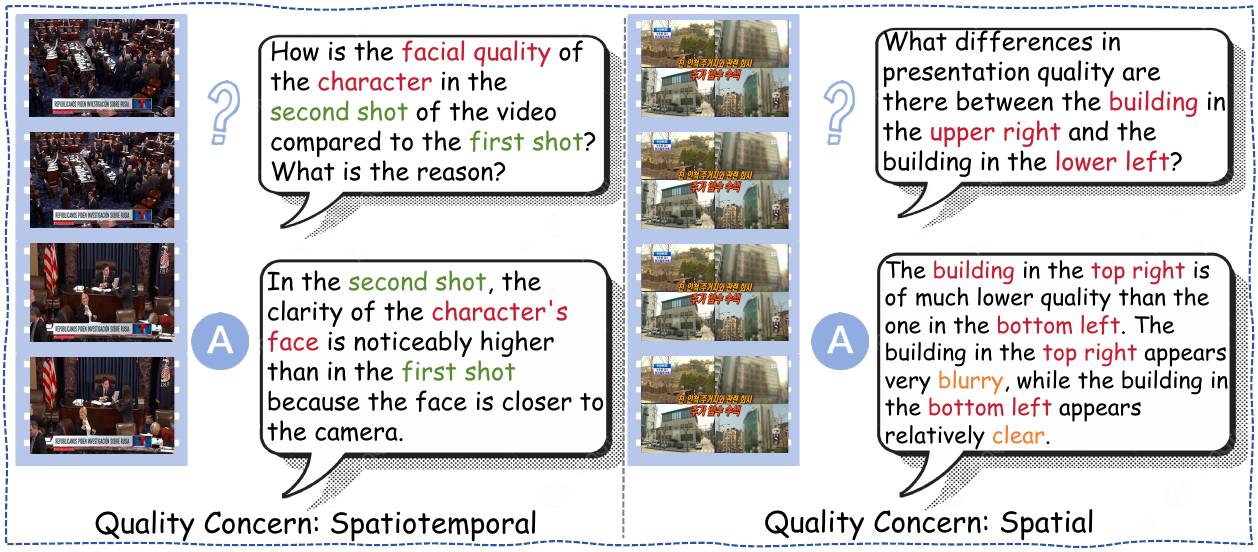}
    \vspace{-8pt}
   \caption{Examples of questions and correct answers of the human-annotated part of the \textbf{OmniVQA-FG-Benchmark}.}
   \vspace{-15pt}
   \label{fig:model}
\end{figure}
\vspace{-5pt}
\section{The OmniVQA-MOS-20K}
\vspace{-2pt}
To enhance the model’s performance on quality rating tasks, we construct a large-scale UGC video subjective scoring dataset comprising $20,000$ video segments and over $200,000$ annotated scores. These selected videos are drawn from a distinct candidate pool of $100,000$ videos, separate from the video pool used in Sec. \ref{OmniVQA-Chat-400K}, with the objective quality label for each video annotated using the same procedure. The video selection aims to ensure a well-balanced distribution of objective quality labels. To this end, we first convert the objective quality levels of all candidate videos to be aligned with the MOSs of LSVQ (train) \cite{ying2021patch}. Then, we ensure that the distribution of objective quality levels for the selected videos mirrors that of LSVQ (train). Specifically, for the $5$ objective quality levels (determined by its objective label) —excellent ($80$-$100$), good ($60$-$80$), fair ($40$-$60$), poor ($20$-$40$), and low ($0$-$20$)— the proportion of videos in each quality level must align with that of LSVQ (train). 

During the subjective rating experiment, we implement a hidden reference supervision strategy where the objective quality level for each video serves as the reference but is not visible to the human annotators. Any human score deviating by two or more levels from the reference is rejected, and rescoring is requested. This method effectively minimizes human annotation bias, especially in scenarios with limited expert-level resources. The distribution of human-labeled scores (averaged for each video) is shown in Fig. \ref{fig:statistical}. Finally, we randomly split the dataset into the \textbf{OmniVQA-MOS-20K (train)} and the \textbf{OmniVQA-MOS-20K (test)} with an $80:20$ ratio for subsequent experiments.

Since the model training still focuses on text-based tasks, the videos and quality labels in \textbf{OmniVQA-MOS-20K (train
)} need to be converted into Q\&A pairs. The specific format is as follows: 
\textit{Q: How would you rate the quality of this video?} The answer is the human-labeled quality rating converted into the quality level.

\vspace{-5pt}
\section{The OmniVQA-FG-Benchmark}

To evaluate the model’s performance in fine-grained tasks, we construct the \textbf{QmniVQA-FG-Benchmark}. This bench contains $1,200$ Q\&A pairs and consists of both machine and human annotations. The statistical information about the benchmark is shown in Fig. \ref{fig:benchstatistic}. The entire benchmark focuses on the following $3$ core quality concerns:

\begin{itemize}
    \item \textbf{Spatial (S)}: Spatial quality primarily concerns quality issues within specific local regions, particularly focusing on specific areas or semantic objects in the video. 
    \item \textbf{Temporal (T)}: Temporal quality concentrates on temporal quality issues, especially issues related to specific timepoints or periods.
    \item \textbf{Spatiotemporal (ST)}: Spatiotemporal quality encompasses both spatial and temporal dimensions, focusing on quality issues that arise in specific locations or at semantic objects during specific timepoints or intervals.
\end{itemize}
Detailed examples are shown in the \textit{supp.} (Figs. \ref{fig:benchmark(1)} and \ref{fig:benchmark(2)}).

For the \textbf{machine annotation}, we select $900$ videos with objective quality labels higher than $70$ in the candidate videos pool in Sec. \ref{OmniVQA-Chat-400K} (different from videos for MIDB annotation) and manually synthesize varying spatiotemporal distortion at different locations and periods. The annotations concentrate on fixed-form descriptions to evaluate the models' ability to localize spatiotemporal distortions. The questions in the machine-annotated part are all multi-choice, single-answer questions.

For the \textbf{human annotation}, we select $1,000$ videos in the candidate videos pool with an objective quality score below $70$ to ensure that the videos contain abundant quality issues to be annotated. Unlike machine annotations, human annotations delve deeper into the analysis and semantic descriptions of spatiotemporal local quality. It follows a flexible approach, incorporating both multi-choice questions and open-ended questions, ensuring a comprehensive capture and annotation of video quality issues.

During the annotation processes, we rigorously follow the requirements in \textbf{ITU-R BT.500-15} to ensure the consistency and accuracy of the annotated data.
\begin{table*}[t]\tiny
    \centering
    \renewcommand\arraystretch{1.05}
    \renewcommand\tabcolsep{6pt}
    \belowrulesep=0pt\aboverulesep=0pt

    \caption{Performance on quality rating tasks. The best result is marked in \textcolor{red}{red},
the second best is denoted in \textcolor{blue}{blue}. The \textbf{OmniVQA-MOS-20K(test)} is presented as ``MOS-20K" in short. }
        \vspace{-8pt}
    \resizebox{1\linewidth}{!}
    {\begin{tabular}{l|cc|cc|cc|cc|cc|cc}
 \hline
    \multicolumn{1}{l|}{\textbf{Datasets}}&\multicolumn{2}{c|}{\textbf{LSVQ(1080p)}\cite{ying2021patch}}& 
    \multicolumn{2}{c|}{\textbf{LSVQ(test)}\cite{ying2021patch}}&\multicolumn{2}{c|}{\textbf{LIVE-VQC}\cite{sinno2018large}}&\multicolumn{2}{c|}{\textbf{KoNViD-1k\cite{hosu2017konstanz}}}&\multicolumn{2}{c|}{\textbf{YT-UGC}\cite{wang2019youtube}}&\multicolumn{2}{c}{\textbf{MOS-20K}}\\ 
    \cdashline{1-13}
       \multicolumn{1}{l|}{\textit{\textbf{Metrics}}}&SRCC &PLCC &SRCC &PLCC &SRCC &PLCC &SRCC &PLCC &SRCC &PLCC&SRCC &PLCC \\ \cdashline{1-13}
    
      \textit{Simple-VQA (ACM.MM 2022)}\cite{sun2022deep} &0.760&0.805 &0.870&0.868 &0.755&0.793 &0.826&0.820 &0.850&0.845 &0.813&0.809 \\
      \textit{BVQA (TCSVT 2022)}\cite{li2022blindly} &0.747&0.785 &0.870&0.861 &0.795&0.814&0.795&0.817&0.845&0.847&0.825&0.813 \\
      \textit{FAST-VQA (TPAMI 2023)}\cite{wu2023neighbourhood} &0.765&0.793 &0.880&0.871 &\textcolor{blue}{0.830}&0.822 &0.869&0.870 &0.828&0.849 &0.792&0.783  \\
     \textit{Dover (ICCV 2023)}\cite{wu2023exploring}&0.797&0.821&0.893&0.892&\textcolor{red}{0.835}&\textcolor{red}{0.857}&0.885&0.879&0.855&0.861&0.828&0.832 \\
     \textit{Modular-VQA (CVPR 2024)}\cite{wen2024modular} &0.810&\textcolor{blue}{0.834}&\textcolor{blue}{0.897}&\textcolor{blue}{0.895}&0.803&0.839&0.876&\textcolor{blue}{0.887}&0.862&\textcolor{red}{0.878}&\textcolor{red}{0.843}&\textcolor{blue}{0.835} \\
    \cdashline{1-13}
    \textit{q-align-VQA (\textit{7B}) (ICML 2024)}\cite{wu2024q1}&0.758 &0.833 &0.883 &0.882 &0.777 &0.813 &0.865 &0.876&0.811 &0.830 &0.820 &0.831\\ 
    \textit{q-align-onealign (\textit{7B})}&0.803&0.836&0.888&0.885&0.773&0.829&0.876 &0.878 &0.831 &0.847 &0.829 &0.826\\ 
   
     \textit{VQA\textsuperscript{\scalebox{0.7}{2}}-UGC-Scorer (\textit{7B})}&0.782&0.837 &0.897&0.885 &0.798&0.830  &\textcolor{blue}{0.894}&0.884 &0.818&0.827 &0.785&0.773 \\
     \cdashline{1-13}
       \textbf{\textit{Chatter (\textit{7B}) (400K)}}&\textcolor{red}{0.816}&0.821&0.889&0.856 &0.822&0.846&0.882&0.835 &\textcolor{blue}{0.859}&0.839&0.810&0.788 \\
      \textbf{\textit{Rater (\textit{7B}) }}&\textcolor{blue}{0.815}&\textcolor{red}{0.838}&\textcolor{red}{0.902}&\textcolor{red}{0.905} &0.826&\textcolor{blue}{0.855}&\textcolor{red}{0.895}&\textcolor{red}{0.900} &0.872&\textcolor{blue}{0.873}&\textcolor{blue}{0.837}&\textcolor{red}{0.837} \\
  \hline
    \end{tabular}}
      \vspace{-2pt}
    \label{tab:rating}
\end{table*}
\begin{table*}[t]\fontsize{6pt}{7pt}\selectfont
    \centering
    \renewcommand\arraystretch{1.12}
    \renewcommand\tabcolsep{3pt}
    \belowrulesep=0pt\aboverulesep=0pt

    \caption{Evaluation results on the {\textit{test}} and {\textit{dev}} subset of the \textit{Q-bench-video}.}
 \vspace{-8pt}
    \resizebox{\linewidth}{!}{\begin{tabular}{l|ccccccc|c|ccccccc|c}
     \hline
     \textbf{Categories} & \multicolumn{8}{|c|}{\textbf{\textit{Q-bench-video-test ($900$ questions)}}} & \multicolumn{8}{c}{\textbf{\textit{Q-bench-video-dev ($892$ questions)}}} 
        \\ \cdashline{1-17}
        \multirow{1}{*}{\textit{LMMs}} & \textit{Binary}&\textit{Multi.} & \textit{Open} & \textit{Tech.}&\textit{Aes.}& \textit{Temp.}  &\textit{AIGC} &\textit{Overall}& \textit{Binary}&\textit{Multi.} & {\textit{Open}} & \textit{Tech.}&\textit{Aes.}& \textit{Temp.}  &\textit{AIGC} &\textit{Overall} \\
      \hline
      \multirow{1}{*}{\textit{mplug-owl3 (\textit{7B})}} \cite{ye2024mplug}&56.90\% & 57.14\% & 42.88\% & 53.40\% & 61.85\% & 50.34\% & 45.34\% & 52.06\%&57.14\% & 54.57\% & 40.37\% & 52.80\% & 58.37\% & 55.09\% & 42.38\% & 51.12\%\\ 
      \multirow{1}{*}{\textit{Internvl2 (\textit{8B})}} \cite{chen2024expanding}&48.15\% & 39.37\% & 31.49\% & 39.06\% & 46.68\% & 42.52\% & 31.37\% & 39.50\% &44.22\% & 31.40\% & 30.56\% & 33.70\% & 49.75\% & 43.27\% & 23.78\% & 35.37\%\\ 
      \multirow{1}{*}{\textit{Internvl2 (\textit{40B})}}&52.53\% & 43.21\% & 35.13\% & 42.54\% & 52.13\% & 46.43\% & 42.55\% & 43.44\%&52.38\% & 33.23\% & 34.63\% & 40.75\% & 53.94\% & 46.00\% & 28.05\% & 39.97\%\\ 
       \multirow{1}{*}{\textit{Internvl2.5 (\textit{8B})}}&46.46\% & 41.81\% & 30.70\% & 39.06\% & 46.21\% & 39.97\% & 28.57\% & 39.44\%&46.26\% & 37.80\% & 31.11\% & 38.29\% & 51.72\% & 44.55\% & 24.70\% & 38.57\%\\ 
      \multirow{1}{*}{\textit{LLaVA-onevision (\textit{7B})}} \cite{li2024llava}&57.58\% & 48.78\% & 32.12\% & 44.98\% & 50.95\% & 45.07\% & 44.72\% & 45.83\% &61.22\% & 50.30\% & 34.63\% & 49.58\% & 59.61\% & 45.64\% & 46.04\% & 49.16\%\\ 
      \multirow{1}{*}{\textit{LLaVA-onevision (\textit{72B})}}&52.19\% & 54.36\% & 34.34\% & 45.62\% & 54.74\% & 50.00\% & 46.58\% & 46.61\% & 54.08\% & 48.78\% & 32.59\% & 47.37\% & 57.39\% & 47.45\% & 30.79\% & 45.63\%\\ 
      \multirow{1}{*}{\textit{Qwen2-vl (\textit{7B})}} \cite{wang2024qwen2}&50.84\% & 55.75\% & 34.49\% & 46.03\% & 56.40\% & 50.17\% & 39.75\% & 46.67\% & 57.48\% & 51.83\% & 32.78\% & 51.19\% & 55.42\% & 51.64\% & 30.49\% & 47.93\%\\ 
      \multirow{1}{*}{\textit{Qwen2-vl (\textit{72B})}}&61.62\% & \textcolor{red}{66.90\%} & 39.24\% & 55.19\% & \textcolor{blue}{63.03\%} & 52.38\% & 50.93\% & 55.44\% &69.73\% & 64.63\% & 38.89\% & 60.87\% & \textcolor{blue}{62.81\%} & 60.18\% & 39.63\% & \textcolor{blue}{58.52\%}\\ 
      \multirow{1}{*}{\textit{Qwen2.5-vl (\textit{7B})}} \cite{bai2025qwen2}&52.53\% & 49.48\% & 38.77\% & 46.68\% & 58.53\% & 46.09\% & 41.61\% & 46.72\%&56.80\% & 45.43\% & 39.26\% & 48.30\% & 60.59\% & 51.27\% & 34.76\% & 47.31\%\\ 
      \multirow{1}{*}{\textit{Qwen2.5-vl (\textit{72B})}}&54.55\% & 49.48\% & 35.92\% & 46.03\% & 57.82\% & 47.96\% & 40.68\% & 46.39\% & 60.20\% & 48.17\% & 35.19\% & 47.62\% & 57.88\% & 54.73\% & 34.45\% & 48.21\%\\ 
      \cdashline{1-17}
      \multirow{1}{*}{\textit{GPT-4o (24-11-20)}} \cite{achiam2023gpt}&60.61\% & 50.17\% & \textcolor{red}{45.25\%} & 50.89\% & \textcolor{red}{63.27\%} & 52.04\% & 48.45\% & 51.89\% &69.73\% & 48.48\% & \textcolor{red}{41.11\%} & 51.27\% & \textcolor{red}{63.55\%} & 59.09\% & 42.68\% & 53.25\%\\ 
      \multirow{1}{*}{\textit{Gemini-1.5-pro}} \cite{team2024gemini}&56.80\% & 43.29\% & 39.26\% & 44.57\% & 53.94\% & 54.64\% & 44.21\% & 46.52\%& 56.80\% & 43.90\% & 37.59\% & 44.14\% & 53.20\% & 54.00\% & \textcolor{blue}{45.73\%} & 46.24\%\\ 
      \multirow{1}{*}{\textit{Gemini-2.0-flash}}&56.23\% & 46.34\% & 43.57\% & 47.73\% & 58.77\% & 48.64\% & \textcolor{red}{55.90\%} & 49.33\% & 64.63\% & 47.26\% & \textcolor{blue}{40.93\%} & 48.64\% & 62.07\% & 58.91\% & \textcolor{red}{51.22\%} & 51.07\%\\ 
      \cdashline{1-17}
      \multirow{1}{*}{\textit{VQA\textsuperscript{2}-Assistant 
 (\textit{7B})}} \cite{jia2024vqa}&\textcolor{blue}{67.12\%} & 59.93\% & 39.56\% & \textcolor{blue}{55.19\%} & 56.87\% & \textcolor{red}{57.99\%} & 43.79\% &  \textcolor{blue}{55.56\%} & \textcolor{blue}{73.81\%} & \textcolor{blue}{56.40\%} & 38.33\% & \textcolor{blue}{60.70\%} & 56.65\% & \textcolor{blue}{61.09\%} & 38.11\% & 56.67\%\\ 
      \multirow{1}{*}{\textbf{\textit{Chatter (\textit{7B}) (400K)}}}&\textcolor{red}{68.35\%} & \textcolor{blue}{63.76\%} & \textcolor{blue}{44.46\%} & \textcolor{red}{58.10\%} & 60.66\% & \textcolor{blue}{54.93\%} & \textcolor{blue}{52.17\%} & \textcolor{red}{58.50\%}& \textcolor{red}{75.51\%} & \textcolor{red}{59.76\%} & 40.37\% & \textcolor{red}{62.05\%} & 61.58\% & \textcolor{red}{63.45\%} & 42.38\% & \textcolor{red}{59.08\%}\\ 
       \hline
      \end{tabular}}
      \vspace{-10pt}

    \label{tab:overall}
\end{table*}

\begin{table*}[t]\tiny
    \centering
    \renewcommand\arraystretch{1.13}
    \renewcommand\tabcolsep{4pt}
    \belowrulesep=0pt\aboverulesep=0pt

\caption{Evaluation results on the OmniVQA-FG-Bench, where ``Mach." denotes ``Machine", ``S" denotes ``Spatial", ``T" denotes ``Temporal", and ``ST" denotes "spatiotemporal".}
    \vspace{-8pt}
    \resizebox{\linewidth}{!}{\begin{tabular}{l|ccc|ccc|c|ccc|c|c}
     \hline
     \textbf{Categories} & \multicolumn{3}{|c|}{\textbf{\textit{Question Types}}} & \multicolumn{4}{c|}{\textbf{\textit{Machine Annotated}}} & \multicolumn{4}{c|}{\textbf{\textit{Human Annotated}}} &\multirow{2}{*}{\textbf{\textit{Overall}}}
        \\ \cdashline{1-12}
        \multirow{1}{*}{\textit{LMMs}} & \textit{Binary}&\textit{Multi.} & \textit{Open} & \textit{Mach. S}&\textit{Mach. T}& \textit{Mach. ST}& \textit{Overall}  &\textit{Human S} &\textit{Human T}& \textit{Human ST} & \textit{Overall}\\
      \hline
      \multirow{1}{*}{\textit{mplug-owl3 (\textit{7B})}}&25.05\% & 30.93\% & 34.74\% & 21.50\% & 17.50\% & 40.00\% & 24.13\% & 37.05\% & 41.82\% & 41.18\% & 39.50\% & 28.88\%\\ 
      \multirow{1}{*}{\textit{Internvl2 (\textit{8B})} }&26.59\% & 24.29\% & 24.03\% & 18.00\% & 15.50\% & 16.00\% & 16.93\% & 30.22\% & 32.27\% & 50.00\% & 34.33\% & 25.25\%\\ 
      \multirow{1}{*}{\textit{Internvl2 (\textit{40B})}}&49.13\% & 31.31\% & 30.19\% & 38.75\% & 34.00\% & 28.00\% & 35.33\% & 39.93\% & 39.55\% & 40.20\% & 39.83\% & 38.88\%\\ 
       \multirow{1}{*}{\textit{Internvl2.5 (\textit{8B})}}&21.97\% & 37.38\% & 21.10\% & 29.75\% & 14.50\% & 28.00\% & 25.33\% & 29.14\% & 27.73\% & 30.39\% & 28.83\% & 28.62\%\\ 
      \multirow{1}{*}{\textit{LLaVA-onevision (\textit{7B})}}&31.60\% & 29.98\% & 27.60\% & 25.00\% & 17.00\% & 33.33\% & 24.53\% & 33.45\% & 40.00\% & 37.25\% & 36.50\% & 30.38\%\\ 
      \multirow{1}{*}{\textit{LLaVA-onevision (\textit{72B})}}&36.03\% & 30.36\% & 24.03\% & 30.75\% & 22.50\% & 33.33\% & 29.07\% & 30.94\% & 38.64\% & 40.20\% & 35.33\% & 32.00\%\\ 
      \multirow{1}{*}{\textit{Qwen2-vl (\textit{7B})}}&26.59\% & 23.15\% & 22.73\% & 20.25\% & 18.50\% & 24.67\% & 20.67\% & 26.26\% & 33.64\% & 42.16\% & 31.67\% & 24.58\%\\ 
      \multirow{1}{*}{\textit{Qwen2-vl (\textit{72B})}}&33.14\% & 40.80\% & 27.27\% & 32.50\% & 26.00\% & 34.00\% & 31.07\% & 39.21\% & 44.55\% & 42.16\% & 41.67\% & 35.75\%\\ 
      \multirow{1}{*}{\textit{Qwen2.5-vl (\textit{7B})}}&23.12\% & 24.29\% & 30.84\% & 16.50\% & 5.50\% & 21.33\% & 14.53\% & 33.09\% & 41.36\% & 43.14\% & 37.83\% & 24.62\%\\ 
      \multirow{1}{*}{\textit{Qwen2.5-vl (\textit{72B})}}&26.20\% & 24.29\% & 24.84\% & 19.00\% & 7.50\% & 25.33\% & 17.20\% & 32.53\% & 43.12\% & 32.61\% & 36.38\% & 25.19\%\\ 
    
      \multirow{1}{*}{\textit{GPT-4o (24-11-20)}}&48.55\% & \textcolor{blue}{39.47\%} & 33.12\% & 48.00\% & 37.00\% & 44.67\% & 44.40\% & 38.85\% & 45.91\% & 42.16\% & 42.00\% & 42.58\%\\ 
        
      \multirow{1}{*}{\textit{VQA\textsuperscript{\scalebox{0.7}{2}}-Assistant (\textit{7B})}}&\textcolor{blue}{76.11\%} & 37.19\% &  \textcolor{blue}{37.99\%} &  \textcolor{blue}{51.25\%} &  \textcolor{blue}{60.00\%} &  \textcolor{blue}{60.00\%} &  \textcolor{blue}{55.33\%} &  \textcolor{blue}{43.88\%} &  \textcolor{red}{50.45\%} &  \textcolor{red}{45.10\%} &  \textcolor{blue}{46.50\%} & \textcolor{blue}{54.12\%}\\ 
      \cdashline{1-13}
      \multirow{1}{*}{\textbf{\textit{Chatter (\textit{7B}) (400K)}}}&\textcolor{red}{81.31\%} & \textcolor{red}{57.31\%} & \textcolor{red}{39.03\%} & \textcolor{red}{56.75\%} & \textcolor{red}{85.50\%} & \textcolor{red}{84.00\%} & \textcolor{red}{69.87\%} & \textcolor{red}{44.52\%} & \textcolor{blue}{46.33\%} &\textcolor{red}{56.52\%} & \textcolor{red}{47.01\%} & \textcolor{red}{65.32\%}\\ 
       \hline
    \end{tabular}}
      
    \vspace{-5pt}
    \label{tab:FG}
\end{table*}
\vspace{-9pt}

\begin{figure}[t]
  \centering
  \includegraphics[width=\linewidth]{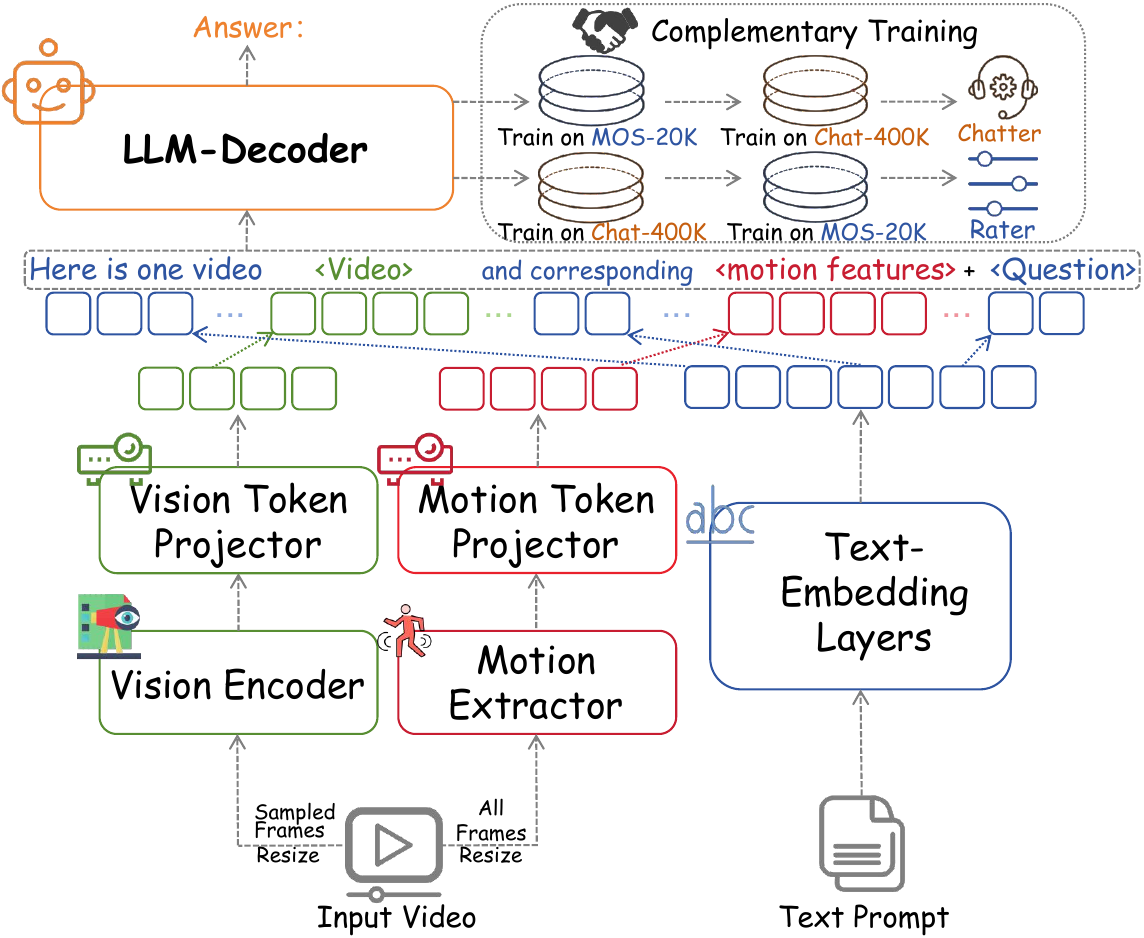}
    \vspace{-8pt}
   \caption{Illustration of OmniVQA models and the complementary training strategy.}
   \vspace{-14pt}
   \label{fig:model}
\end{figure}

\vspace{8pt}
\section{The OmniVQA Models}

After obtaining the OmniVQA-Chat-400K and OmniVQA-MOS-20K MIDBs, the \textit{VQA\textsuperscript{2}-Stage-1} model \cite{jia2024vqa} is employed as the base model for supervised fine-tuning (SFT). The base model consists of the \textit{SigLip} \cite{zhai2023sigmoid} vision encoder, the \textit{SlowFast-R50} \cite{feichtenhofer2019slowfast} motion extractor, and the \textit{Qwen-2} \cite{yang2024qwen2} LLM model (shown in Fig. \ref{fig:model}). In this model, the text tokens from the prompt, the vision tokens from the video keyframes, and the motion tokens from the entire video are \textbf{interleaved} into a semantically ordered sequence, which is then input into the LLM for text generation.
Through different SFT processes, we obtain $2$ specialized models: The \textbf{\textit{rater}} focuses on perceptual VQA quantification rating tasks, and the \textbf{\textit{chatter}} specializes in quality understanding and question-answering tasks.

\paragraph{Complementary training strategy.}
We posit that the role of the LLM part varies between quantitative rating tasks and quality understanding tasks.  In the former, the LLM functions primarily as an effective \textbf{regressor}, whereas in the latter, it learns to navigate the complex semantic relationships between different quality factors and modalities. Consequently, we argue that random mixing of training data from these tasks may undermine the LLM's ability to effectively perform on each task, as the divergent training objectives could hinder its capacity to focus on the specific learning goals. However, from a pre-training perspective, these two tasks are perfectly complementary. Firstly, the sequential training process mitigates potential confusion regarding the model's learning objectives. More importantly, the intrinsic relationship between the knowledge of the two tasks suggests that the datasets can provide valuable \textbf{prior information} for one another, thus making them well-suited to serve as mutually beneficial pre-training components. The process of complementary training is also depicted in Fig. \ref{fig:model}.
\label{sec:model}

\vspace{-18pt}
\section{Experiments}
\vspace{-6pt}
We conduct a detailed evaluation of our models on video quality rating and video quality understanding tasks. In addition, we perform comprehensive supplementary experiments to investigate some key factors.
\vspace{-5pt}
\subsection{Experimental Setups}
\paragraph{System prompt design.}
In all evaluations, we set almost unified system prompts for all LMM models; the system prompts are illustrated in the \textit{supp.} (Sec. \ref{system prompts}).

\paragraph{Model training.}
We employ the complementary training strategy to obtain the models. All training is performed with full-parameter-tuning, with only $1$ epoch trained on each dataset. The specific hyper-parameter configurations and model structures are also presented in the \textit{supp.} (Tab. \ref{tab:hyperparam}). 
\vspace{-8pt}
\subsection{Evaluation on quality rating tasks}
\vspace{-5pt}
We compare our models with several DNN-based \cite{sun2022deep,li2022blindly,wu2023neighbourhood,wu2023exploring,wen2024modular}
and LMM-based  \cite{wu2024q1,jia2024vqa} quality rating models on $6$ datasets including the \textbf{OmniVQA-MOS-20K (test)}. Apart from the \textit{Q-align}, \textit{VQA\textsuperscript{2}-UGC-Scorer}, and our models (using complementary training), all models are trained on the merged dataset of OmniVQA-MOS-20K (train) and LSVQ (train) (approximately $43,000$ videos). The evaluation metrics are the \textit{Pearson Linear Correlation Coefficient} (PLCC) and  \textit{Spearman Rank Correlation Coefficient} (SRCC). We adopt the quality rating method used in \cite{wu2024q1,jia2024vqa} during testing, which is detailed in the \textit{supp.} (Sec. \ref{evaluation}).
The performance of the models on all datasets is presented in 
Tab.\ref{tab:rating}. The experimental results demonstrate that the \textbf{rater} achieves \textit{Top-3} performance across all $6$ datasets. This demonstrates its superior performance in quality rating tasks. Since the \textbf{chatter} model is not a proprietary model for the rating task, its performance shows a noticeable decline, but it still indicates acceptable performance.

\begin{table*}[t]\tiny
    \centering
    \renewcommand\arraystretch{1.13}
    \renewcommand\tabcolsep{5pt}
    \belowrulesep=0pt\aboverulesep=0pt

    \caption{Performance of training strategies. The \textbf{Overall} task is evaluated on the \textit{q-bench-video (test)}. The best result is denoted in \textcolor{red}{red}.}
    \vspace{-9pt}
    \resizebox{1\linewidth}{!}
    {\begin{tabular}{l|c|c|c|c|c|c|c|c|c|c|c}
 \hline
    \textbf{Categories}&\multicolumn{5}{c|}{\textbf{Quality Rating (SRCC~/~PLCC)}}&\multicolumn{3}{c|}{\textbf{Overall}}&\multicolumn{3}{c}{\textbf{Fine-grain}}\\ 
     \cdashline{1-12}
    \textbf{Training Strategy}&\textit{LSVQ(1080p)}&\textit{LSVQ(test)}&\textit{LIVE-VQC}&\textit{KoNViD-1k}&\textit{OmniVQA (test)}&\textit{Tech.}&\textit{Aes.}&\textit{Overall}&\textit{Machine}&\textit{Human}&\textit{Overall}\\
   \cdashline{1-12}
      \textit{Direct}&0.800~/~0.824&0.880~/~0.878&0.776~/~0.820&0.877~/~0.883&0.819~/~0.820 &57.54\% &\textcolor{red}{63.03\%} &58.33\% &68.00\% &48.50\% &62.28\% \\
      \textit{Mix}&\textcolor{red}{0.817}~/~0.836&0.898~/~0.896&0.822~/~\textcolor{red}{0.860}&0.887~/~0.898&\textcolor{red}{0.840}~/~\textcolor{red}{0.838} &52.54\%&56.32\%&52.78\% &66.47\%&51.17\%&63.58\%\\
      \textbf{\textit{Complementary}}&0.815~/~\textcolor{red}{0.838}&\textcolor{red}{0.902}~/~\textcolor{red}{0.905}&\textcolor{red}{0.826}~/~0.855&\textcolor{red}{0.895}~/~\textcolor{red}{0.900}&0.837~/~0.837&\textcolor{red}{58.10\%}&60.66\%&\textcolor{red}{58.50\%}&\textcolor{red}{69.87\%}&\textcolor{red}{56.52\%}&\textcolor{red}{65.32\%} \\
  \hline
    \end{tabular}}
  \vspace{-9pt}
    \label{tab:trainingablation}
\end{table*}

\subsection{Evaluation on quality understanding tasks}
As the primary task of our work, we conduct detailed video quality understanding evaluation experiments, which include both overall and fine-grained tasks. We also include $4$ \textbf{real-world scenario case studies} in the \textit{supp.} (Figs. \ref{fig:case1}, \ref{fig:case2}, \ref{fig:case3} and \ref{fig:case4}) to visualize the functionality of the model.

The overall video quality understanding task is carried out on the \textbf{Q-bench-video}. Since our training process does not include multi-video comparison analysis, we remove questions involving multi-video quality issues. The evaluation question types include binary questions (\textit{Binary}), multi-choice (single-answer) questions (\textit{Multi.}), and open-ended questions (\textit{Open}). The questions cover different quality concerns, including technical quality (\textit{Tech.}), aesthetic quality (\textit{Aes.}), temporal quality (\textit{Temp.}), and AIGC video quality (\textit{AIGC}). We select some of the latest open-source LMMs with video analysis capabilities with varying parameter sizes, some proprietary LMMs, and the \textit{VQA\textsuperscript{2}-Assistant} for comparison. To ensure a fair comparison, the input for each model is only related to its architecture. For models without a motion extractor, we input the keyframe sequence obtained by sampling $1$ frame per second from the video. For the \textit{VQA\textsuperscript{2}-Assistant} and our chatter model, we additionally input the whole frame sequence of the video (resizing to $(224*224)$ to the motion extractor. The experimental results are presented in Tab. \ref{tab:overall}.


The experimental results show that the \textbf{chatter} achieves the best \textit{Overall} performance on both the \textit{test} and \textit{dev} subsets, with outstanding performance in the \textit{Tech., Aes., and Temp.} quality concerns. Although it does not outperform some of the most advanced LMMs on some subcategories, the performance gap is relatively minimal. This demonstrates that in overall video quality understanding tasks, synthetic MIDBs with human-in-the-loop can still yield superior training performance.

For the fine-grained video quality understanding task, we conduct comparison experiments between the \textbf{chatter} model and the comparison models on the \textbf{OmniVQA-FG-Bench}. Tab. \ref{tab:FG} records the performance of each LMM on different subcategories of questions in the benchmark. Models specially trained on VQA tasks (\textit{VQA\textsuperscript{2}-Assistant (7B)} and \textbf{chatter}) achieve significantly superior performance compared to general LMMs. Additionally, our \textbf{chatter} outperforms \textit{VQA\textsuperscript{2}-Assistant} by a significant margin on machine annotation tasks and also achieves better performance on human annotation tasks. This demonstrates the importance of the in-context branch in the MIDB for improving the model's performance in fine-grained spatiotemporal quality understanding tasks.

\vspace{-1pt}
\subsection{Discussions}
\vspace{-1pt}
We have also conducted experiments on some key points, and further analysis is presented in the \textit{supp.} (Sec. \ref{analysis}).

\begin{figure}[t]
  \centering
  \includegraphics[width=0.95\linewidth]{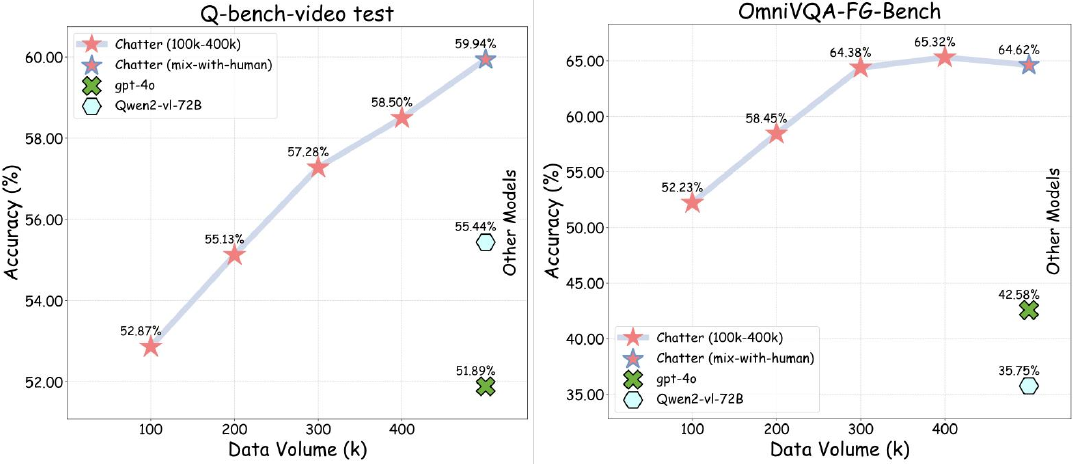}
    \vspace{-5pt}
   \caption{The performance map of scaling up the training dataset.}
   \vspace{-13pt}
   \label{fig:datascaling}
\end{figure}

\paragraph{Data scaling effects verification.}
Additionally, we validate the data-scaling effect by selecting subsets of data (ranging from $100$k to $400$k, with each subset having an equal distribution across $3$ branches) from OmniVQA-Chat-400K for ablation. The performance is shown in Fig. \ref{fig:datascaling}, where \textit{`mix with human'} denotes that we combine the OmniVQA-Chat-400K dataset with the \textit{VQA\textsuperscript{2}-stage3}\cite{jia2024vqa} (which contains $115$K human-annotated, \textit{sota LMM}-refined high-quality data) and then retrain the base model.  It is evident that the data-scaling effect appears in the $100$k-$400$k range but gradually becomes marginal under $7$B parameter size. While mixed training with human-annotated data further improves the performance on the \textit{Q-bench-video (test)}, it does not showcase a significant effect on fine-grain tasks.

\paragraph{Effects of complementary training.}
We verify the effects of the complementary training strategy by comparing $2$ other training strategies: the \textit{Direct} strategy, where the model is trained directly on the corresponding dataset for each task, and the \textit{Mix} strategy in which the $2$ datasets are randomly mixed to train one unified model. The experimental results are presented in Tab. \ref{tab:trainingablation}. The results show that our \textit{Complementary} strategy outperforms the \textit{Direct} strategy on all tasks. While the \textit{Mix} strategy shows no significant difference from the \textit{Complementary} strategy in the quality rating task, it exhibits a clear performance gap in the quality understanding tasks. The results showcase the rationale of the complementary training strategy.


\vspace{-3pt}
\section{Conclusion}
In \textbf{Q-Bench-Video}, our \textbf{chatter} model achieves superior performance, highlighting its proficiency in handling general VQA quality understanding tasks. Additionally, it excels on the OmniVQA-FG-Benchmark, showcasing its capability for fine-grained evaluation. Furthermore, on quality rating tasks, our \textbf{rater} model achieves state-of-the-art performance, demonstrating its effectiveness. These results highlight the potential of scaling up perceptual video quality assessment through the integration of machine annotation strategies and task-specific complementary model training.

\newpage
{
    \small
    \bibliographystyle{ieeenat_fullname}
    \bibliography{main}

\begin{thebibliography}{59}
\providecommand{\natexlab}[1]{#1}
\providecommand{\url}[1]{\texttt{#1}}
\expandafter\ifx\csname urlstyle\endcsname\relax
  \providecommand{\doi}[1]{doi: #1}\else
  \providecommand{\doi}{doi: \begingroup \urlstyle{rm}\Url}\fi

\bibitem[Achiam et~al.(2023)Achiam, Adler, Agarwal, Ahmad, Akkaya, Aleman,
  Almeida, Altenschmidt, Altman, Anadkat, et~al.]{achiam2023gpt}
Josh Achiam, Steven Adler, Sandhini Agarwal, Lama Ahmad, Ilge Akkaya,
  Florencia~Leoni Aleman, Diogo Almeida, Janko Altenschmidt, Sam Altman,
  Shyamal Anadkat, et~al.
\newblock Gpt-4 technical report.
\newblock \emph{arXiv preprint arXiv:2303.08774}, 2023.

\bibitem[Bai et~al.(2025)Bai, Chen, Liu, Wang, Ge, Song, Dang, Wang, Wang,
  Tang, et~al.]{bai2025qwen2}
Shuai Bai, Keqin Chen, Xuejing Liu, Jialin Wang, Wenbin Ge, Sibo Song, Kai
  Dang, Peng Wang, Shijie Wang, Jun Tang, et~al.
\newblock Qwen2. 5-vl technical report.
\newblock \emph{arXiv preprint arXiv:2502.13923}, 2025.

\bibitem[Bampis and Bovik(2018)]{bampis2018feature}
Christos~G Bampis and Alan~C Bovik.
\newblock Feature-based prediction of streaming video qoe: Distortions,
  stalling and memory.
\newblock \emph{SPIC}, 68:\penalty0 218--228, 2018.

\bibitem[Bampis et~al.(2021)Bampis, Li, Katsavounidis, Huang, Ekanadham, and
  Bovik]{bampis2021towards}
Christos~G Bampis, Zhi Li, Ioannis Katsavounidis, Te-Yuan Huang, Chaitanya
  Ekanadham, and Alan~C Bovik.
\newblock Towards perceptually optimized adaptive video streaming-a realistic
  quality of experience database.
\newblock \emph{IEEE TIP}, 30:\penalty0 5182--5197, 2021.

\bibitem[Chen et~al.(2024{\natexlab{a}})Chen, Yang, Wu, Liao, Zhang, Wang, Sun,
  Yan, and Lin]{chen2024q}
Chaofeng Chen, Sensen Yang, Haoning Wu, Liang Liao, Zicheng Zhang, Annan Wang,
  Wenxiu Sun, Qiong Yan, and Weisi Lin.
\newblock Q-ground: Image quality grounding with large multi-modality models.
\newblock In \emph{ACM MM}, pages 486--495, 2024{\natexlab{a}}.

\bibitem[Chen et~al.(2024{\natexlab{b}})Chen, Siarohin, Menapace, Deyneka,
  Chao, Jeon, Fang, Lee, Ren, Yang, et~al.]{chen2024panda}
Tsai-Shien Chen, Aliaksandr Siarohin, Willi Menapace, Ekaterina Deyneka,
  Hsiang-wei Chao, Byung~Eun Jeon, Yuwei Fang, Hsin-Ying Lee, Jian Ren,
  Ming-Hsuan Yang, et~al.
\newblock Panda-70m: Captioning 70m videos with multiple cross-modality
  teachers.
\newblock In \emph{CVPR}, pages 13320--13331, 2024{\natexlab{b}}.

\bibitem[Chen et~al.(2024{\natexlab{c}})Chen, Wang, Cao, Liu, Gao, Cui, Zhu,
  Ye, Tian, Liu, et~al.]{chen2024expanding}
Zhe Chen, Weiyun Wang, Yue Cao, Yangzhou Liu, Zhangwei Gao, Erfei Cui, Jinguo
  Zhu, Shenglong Ye, Hao Tian, Zhaoyang Liu, et~al.
\newblock Expanding performance boundaries of open-source multimodal models
  with model, data, and test-time scaling.
\newblock \emph{arXiv preprint arXiv:2412.05271}, 2024{\natexlab{c}}.

\bibitem[Chen et~al.(2024{\natexlab{d}})Chen, Zhang, Li, Pei, Song, Min, Liu,
  Yuan, Guo, and Zhang]{chen2024grounding}
Zheng Chen, Xun Zhang, Wenbo Li, Renjing Pei, Fenglong Song, Xiongkuo Min,
  Xiaohong Liu, Xin Yuan, Yong Guo, and Yulun Zhang.
\newblock Grounding-iqa: Multimodal language grounding model for image quality
  assessment.
\newblock \emph{arXiv preprint arXiv:2411.17237}, 2024{\natexlab{d}}.

\bibitem[Duanmu et~al.(2016)Duanmu, Zeng, Ma, Rehman, and
  Wang]{duanmu2016quality}
Zhengfang Duanmu, Kai Zeng, Kede Ma, Abdul Rehman, and Zhou Wang.
\newblock A quality-of-experience index for streaming video.
\newblock \emph{IEEE JSTSP}, 11\penalty0 (1):\penalty0 154--166, 2016.

\bibitem[Duanmu et~al.(2018)Duanmu, Rehman, and Wang]{duanmu2018quality}
Zhengfang Duanmu, Abdul Rehman, and Zhou Wang.
\newblock A quality-of-experience database for adaptive video streaming.
\newblock \emph{IEEE TBC}, 64\penalty0 (2):\penalty0 474--487, 2018.

\bibitem[Duanmu et~al.(2023)Duanmu, Liu, Chen, Li, Wang, Wang, and
  Gao]{duanmu2023bayesian}
Zhengfang Duanmu, Wentao Liu, Diqi Chen, Zhuoran Li, Zhou Wang, Yizhou Wang,
  and Wen Gao.
\newblock A bayesian quality-of-experience model for adaptive streaming videos.
\newblock \emph{ACM TOMM}, 18\penalty0 (3s):\penalty0 1--24, 2023.

\bibitem[Feichtenhofer et~al.(2019)Feichtenhofer, Fan, Malik, and
  He]{feichtenhofer2019slowfast}
Christoph Feichtenhofer, Haoqi Fan, Jitendra Malik, and Kaiming He.
\newblock Slowfast networks for video recognition.
\newblock In \emph{ICCV}, pages 6202--6211, 2019.

\bibitem[Ge et~al.(2024)Ge, Sun, Zhang, Li, Ji, Sun, Jui, Min, and
  Zhai]{ge2024lmm}
Qihang Ge, Wei Sun, Yu Zhang, Yunhao Li, Zhongpeng Ji, Fengyu Sun, Shangling
  Jui, Xiongkuo Min, and Guangtao Zhai.
\newblock Lmm-vqa: Advancing video quality assessment with large multimodal
  models.
\newblock \emph{arXiv preprint arXiv:2408.14008}, 2024.

\bibitem[Ghadiyaram et~al.(2017)Ghadiyaram, Pan, Bovik, Moorthy, Panda, and
  Yang]{ghadiyaram2017capture}
Deepti Ghadiyaram, Janice Pan, Alan~C Bovik, Anush~Krishna Moorthy, Prasanjit
  Panda, and Kai-Chieh Yang.
\newblock In-capture mobile video distortions: A study of subjective behavior
  and objective algorithms.
\newblock \emph{IEEE TCSVT}, 28\penalty0 (9):\penalty0 2061--2077, 2017.

\bibitem[Hosu et~al.(2017)Hosu, Hahn, Jenadeleh, Lin, Men, Szir{\'a}nyi, Li,
  and Saupe]{hosu2017konstanz}
Vlad Hosu, Franz Hahn, Mohsen Jenadeleh, Hanhe Lin, Hui Men, Tam{\'a}s
  Szir{\'a}nyi, Shujun Li, and Dietmar Saupe.
\newblock The konstanz natural video database (konvid-1k).
\newblock In \emph{QoMEX}, pages 1--6. IEEE, 2017.

\bibitem[Huang et~al.(2024)Huang, Sheng, Yang, Yuan, Duan, Chen, Li, Lin, and
  Shi]{huang2024aesexpert}
Yipo Huang, Xiangfei Sheng, Zhichao Yang, Quan Yuan, Zhichao Duan, Pengfei
  Chen, Leida Li, Weisi Lin, and Guangming Shi.
\newblock Aesexpert: Towards multi-modality foundation model for image
  aesthetics perception.
\newblock In \emph{ACM MM}, pages 5911--5920, 2024.

\bibitem[Jia et~al.(2024)Jia, Zhang, Qian, Wu, Sun, Li, Liu, Lin, Zhai, and
  Min]{jia2024vqa}
Ziheng Jia, Zicheng Zhang, Jiaying Qian, Haoning Wu, Wei Sun, Chunyi Li,
  Xiaohong Liu, Weisi Lin, Guangtao Zhai, and Xiongkuo Min.
\newblock Vqa2: Visual question answering for video quality assessment.
\newblock \emph{arXiv preprint arXiv:2411.03795}, 2024.

\bibitem[Khaki et~al.(2024)Khaki, Li, Ma, Yang, and Ramachandra]{khaki2024rs}
Saeed Khaki, JinJin Li, Lan Ma, Liu Yang, and Prathap Ramachandra.
\newblock Rs-dpo: A hybrid rejection sampling and direct preference
  optimization method for alignment of large language models.
\newblock \emph{arXiv preprint arXiv:2402.10038}, 2024.

\bibitem[Korhonen(2019)]{korhonen2019two}
Jari Korhonen.
\newblock Two-level approach for no-reference consumer video quality
  assessment.
\newblock \emph{IEEE TIP}, 28\penalty0 (12):\penalty0 5923--5938, 2019.

\bibitem[Li et~al.(2022)Li, Zhang, Tian, Zhai, and Wang]{li2022blindly}
Bowen Li, Weixia Zhang, Meng Tian, Guangtao Zhai, and Xianpei Wang.
\newblock Blindly assess quality of in-the-wild videos via quality-aware
  pre-training and motion perception.
\newblock \emph{IEEE TCSVT}, 32\penalty0 (9):\penalty0 5944--5958, 2022.

\bibitem[Li et~al.(2024)Li, Zhang, Guo, Zhang, Li, Zhang, Zhang, Zhang, Li,
  Liu, et~al.]{li2024llava}
Bo Li, Yuanhan Zhang, Dong Guo, Renrui Zhang, Feng Li, Hao Zhang, Kaichen
  Zhang, Peiyuan Zhang, Yanwei Li, Ziwei Liu, et~al.
\newblock Llava-onevision: Easy visual task transfer.
\newblock \emph{arXiv preprint arXiv:2408.03326}, 2024.

\bibitem[Li et~al.(2019)Li, Jiang, and Jiang]{li2019quality}
Dingquan Li, Tingting Jiang, and Ming Jiang.
\newblock Quality assessment of in-the-wild videos.
\newblock In \emph{ACM MM}, pages 2351--2359, 2019.

\bibitem[Li et~al.()Li, Duanmu, Liu, and Wang]{li11662comparative}
Z Li, Z Duanmu, W Liu, and Z Wang.
\newblock A comparative study of state-of-the-art video encoders on 4k videos.
\newblock \emph{Image Analysis and Recognition. LNCS}, 11662.

\bibitem[Liu et~al.()Liu, Zhao, Joshi, Khalman, Saleh, Liu, and
  Liu]{liustatistical}
Tianqi Liu, Yao Zhao, Rishabh Joshi, Misha Khalman, Mohammad Saleh, Peter~J
  Liu, and Jialu Liu.
\newblock Statistical rejection sampling improves preference optimization.
\newblock In \emph{ICLR}.

\bibitem[Liu et~al.(2018)Liu, Duanmu, and Wang]{liu2018end}
Wentao Liu, Zhengfang Duanmu, and Zhou Wang.
\newblock End-to-end blind quality assessment of compressed videos using deep
  neural networks.
\newblock In \emph{ACM MM}, pages 546--554, 2018.

\bibitem[Min et~al.(2024)Min, Duan, Sun, Zhu, and Zhai]{min2024perceptual}
Xiongkuo Min, Huiyu Duan, Wei Sun, Yucheng Zhu, and Guangtao Zhai.
\newblock Perceptual video quality assessment: A survey.
\newblock \emph{SCIS}, 67\penalty0 (11):\penalty0 211301, 2024.

\bibitem[Mittal et~al.(2012{\natexlab{a}})Mittal, Moorthy, and
  Bovik]{mittal2012no}
Anish Mittal, Anush~Krishna Moorthy, and Alan~Conrad Bovik.
\newblock No-reference image quality assessment in the spatial domain.
\newblock \emph{IEEE TIP}, 21\penalty0 (12):\penalty0 4695--4708,
  2012{\natexlab{a}}.

\bibitem[Mittal et~al.(2012{\natexlab{b}})Mittal, Soundararajan, and
  Bovik]{mittal2012making}
Anish Mittal, Rajiv Soundararajan, and Alan~C Bovik.
\newblock Making a “completely blind” image quality analyzer.
\newblock \emph{IEEE SPL}, 20\penalty0 (3):\penalty0 209--212,
  2012{\natexlab{b}}.

\bibitem[Nuutinen et~al.(2016)Nuutinen, Virtanen, Vaahteranoksa, Vuori,
  Oittinen, and H{\"a}kkinen]{nuutinen2016cvd2014}
Mikko Nuutinen, Toni Virtanen, Mikko Vaahteranoksa, Tero Vuori, Pirkko
  Oittinen, and Jukka H{\"a}kkinen.
\newblock Cvd2014—a database for evaluating no-reference video quality
  assessment algorithms.
\newblock \emph{IEEE TIP}, 25\penalty0 (7):\penalty0 3073--3086, 2016.

\bibitem[Radford et~al.(2018)Radford, Narasimhan, Salimans, Sutskever,
  et~al.]{radford2018improving}
Alec Radford, Karthik Narasimhan, Tim Salimans, Ilya Sutskever, et~al.
\newblock Improving language understanding by generative pre-training.
\newblock 2018.

\bibitem[Sinno and Bovik(2018)]{sinno2018large}
Zeina Sinno and Alan~Conrad Bovik.
\newblock Large-scale study of perceptual video quality.
\newblock \emph{IEEE TIP}, 28\penalty0 (2):\penalty0 612--627, 2018.

\bibitem[Sun et~al.(2022)Sun, Min, Lu, and Zhai]{sun2022deep}
Wei Sun, Xiongkuo Min, Wei Lu, and Guangtao Zhai.
\newblock A deep learning based no-reference quality assessment model for ugc
  videos.
\newblock In \emph{ACM MM}, pages 856--865, 2022.

\bibitem[Sun et~al.(2024)Sun, Wen, Min, Lan, Zhai, and Ma]{sun2024analysis}
Wei Sun, Wen Wen, Xiongkuo Min, Long Lan, Guangtao Zhai, and Kede Ma.
\newblock Analysis of video quality datasets via design of minimalistic video
  quality models.
\newblock \emph{IEEE TPAMI}, 2024.

\bibitem[Team et~al.(2024)Team, Georgiev, Lei, Burnell, Bai, Gulati, Tanzer,
  Vincent, Pan, Wang, et~al.]{team2024gemini}
Gemini Team, Petko Georgiev, Ving~Ian Lei, Ryan Burnell, Libin Bai, Anmol
  Gulati, Garrett Tanzer, Damien Vincent, Zhufeng Pan, Shibo Wang, et~al.
\newblock Gemini 1.5: Unlocking multimodal understanding across millions of
  tokens of context.
\newblock \emph{arXiv preprint arXiv:2403.05530}, 2024.

\bibitem[Tu et~al.(2021{\natexlab{a}})Tu, Wang, Birkbeck, Adsumilli, and
  Bovik]{tu2021ugc}
Zhengzhong Tu, Yilin Wang, Neil Birkbeck, Balu Adsumilli, and Alan~C Bovik.
\newblock Ugc-vqa: Benchmarking blind video quality assessment for user
  generated content.
\newblock \emph{IEEE TIP}, 30:\penalty0 4449--4464, 2021{\natexlab{a}}.

\bibitem[Tu et~al.(2021{\natexlab{b}})Tu, Yu, Wang, Birkbeck, Adsumilli, and
  Bovik]{tu2021rapique}
Zhengzhong Tu, Xiangxu Yu, Yilin Wang, Neil Birkbeck, Balu Adsumilli, and
  Alan~C Bovik.
\newblock Rapique: Rapid and accurate video quality prediction of user
  generated content.
\newblock \emph{IEEE OJSP}, 2:\penalty0 425--440, 2021{\natexlab{b}}.

\bibitem[Wang et~al.(2024{\natexlab{a}})Wang, Bai, Tan, Wang, Fan, Bai, Chen,
  Liu, Wang, Ge, et~al.]{wang2024qwen2}
Peng Wang, Shuai Bai, Sinan Tan, Shijie Wang, Zhihao Fan, Jinze Bai, Keqin
  Chen, Xuejing Liu, Jialin Wang, Wenbin Ge, et~al.
\newblock Qwen2-vl: Enhancing vision-language model's perception of the world
  at any resolution.
\newblock \emph{arXiv preprint arXiv:2409.12191}, 2024{\natexlab{a}}.

\bibitem[Wang et~al.(2019)Wang, Inguva, and Adsumilli]{wang2019youtube}
Yilin Wang, Sasi Inguva, and Balu Adsumilli.
\newblock Youtube ugc dataset for video compression research.
\newblock In \emph{IEEE MMSP}, pages 1--5. IEEE, 2019.

\bibitem[Wang et~al.(2021)Wang, Ke, Talebi, Yim, Birkbeck, Adsumilli, Milanfar,
  and Yang]{wang2021rich}
Yilin Wang, Junjie Ke, Hossein Talebi, Joong~Gon Yim, Neil Birkbeck, Balu
  Adsumilli, Peyman Milanfar, and Feng Yang.
\newblock Rich features for perceptual quality assessment of ugc videos.
\newblock In \emph{CVPR}, pages 13435--13444, 2021.

\bibitem[Wang et~al.(2024{\natexlab{b}})Wang, Xu, and Ren]{wang2024llm}
Zhenhua Wang, Guang Xu, and Ming Ren.
\newblock Llm-generated natural language meets scaling laws: New explorations
  and data augmentation methods.
\newblock \emph{arXiv preprint arXiv:2407.00322}, 2024{\natexlab{b}}.

\bibitem[Wen et~al.(2024)Wen, Li, Zhang, Liao, Li, Zhang, and
  Ma]{wen2024modular}
Wen Wen, Mu Li, Yabin Zhang, Yiting Liao, Junlin Li, Li Zhang, and Kede Ma.
\newblock Modular blind video quality assessment.
\newblock In \emph{CVPR}, pages 2763--2772, 2024.

\bibitem[Wu et~al.(2023{\natexlab{a}})Wu, Chen, Liao, Hou, Sun, Yan, Gu, and
  Lin]{wu2023neighbourhood}
Haoning Wu, Chaofeng Chen, Liang Liao, Jingwen Hou, Wenxiu Sun, Qiong Yan,
  Jinwei Gu, and Weisi Lin.
\newblock Neighbourhood representative sampling for efficient end-to-end video
  quality assessment.
\newblock \emph{IEEE TPAMI}, 45\penalty0 (12):\penalty0 15185--15202,
  2023{\natexlab{a}}.

\bibitem[Wu et~al.(2023{\natexlab{b}})Wu, Chen, Liao, Hou, Sun, Yan, and
  Lin]{wu2023discovqa}
Haoning Wu, Chaofeng Chen, Liang Liao, Jingwen Hou, Wenxiu Sun, Qiong Yan, and
  Weisi Lin.
\newblock Discovqa: Temporal distortion-content transformers for video quality
  assessment.
\newblock \emph{IEEE TCSVT}, 33\penalty0 (9):\penalty0 4840--4854,
  2023{\natexlab{b}}.

\bibitem[Wu et~al.(2023{\natexlab{c}})Wu, Zhang, Liao, Chen, Hou, Wang, Sun,
  Yan, and Lin]{wu2023exploring}
Haoning Wu, Erli Zhang, Liang Liao, Chaofeng Chen, Jingwen Hou, Annan Wang,
  Wenxiu Sun, Qiong Yan, and Weisi Lin.
\newblock Exploring video quality assessment on user generated contents from
  aesthetic and technical perspectives.
\newblock In \emph{ICCV}, pages 20144--20154, 2023{\natexlab{c}}.

\bibitem[Wu et~al.(2023{\natexlab{d}})Wu, Zhang, Liao, Chen, Hou, Wang, Sun,
  Yan, and Lin]{wu2023towards}
Haoning Wu, Erli Zhang, Liang Liao, Chaofeng Chen, Jingwen Hou, Annan Wang,
  Wenxiu Sun, Qiong Yan, and Weisi Lin.
\newblock Towards explainable in-the-wild video quality assessment: A database
  and a language-prompted approach.
\newblock In \emph{ACM MM}, pages 1045--1054, 2023{\natexlab{d}}.

\bibitem[Wu et~al.(2024{\natexlab{a}})Wu, Zhang, Zhang, Chen, Liao, Wang, Xu,
  Li, Hou, Zhai, et~al.]{wu2024q}
Haoning Wu, Zicheng Zhang, Erli Zhang, Chaofeng Chen, Liang Liao, Annan Wang,
  Kaixin Xu, Chunyi Li, Jingwen Hou, Guangtao Zhai, et~al.
\newblock Q-instruct: Improving low-level visual abilities for multi-modality
  foundation models.
\newblock In \emph{CVPR}, pages 25490--25500, 2024{\natexlab{a}}.

\bibitem[Wu et~al.(2024{\natexlab{b}})Wu, Zhang, Zhang, Chen, Liao, Li, Gao,
  Wang, Zhang, Sun, et~al.]{wu2024q1}
Haoning Wu, Zicheng Zhang, Weixia Zhang, Chaofeng Chen, Liang Liao, Chunyi Li,
  Yixuan Gao, Annan Wang, Erli Zhang, Wenxiu Sun, et~al.
\newblock Q-align: teaching lmms for visual scoring via discrete text-defined
  levels.
\newblock In \emph{ICML}, pages 54015--54029, 2024{\natexlab{b}}.

\bibitem[Wu et~al.(2024{\natexlab{c}})Wu, Zhu, Zhang, Zhang, Chen, Liao, Li,
  Wang, Sun, Yan, et~al.]{wu2024towards}
Haoning Wu, Hanwei Zhu, Zicheng Zhang, Erli Zhang, Chaofeng Chen, Liang Liao,
  Chunyi Li, Annan Wang, Wenxiu Sun, Qiong Yan, et~al.
\newblock Towards open-ended visual quality comparison.
\newblock In \emph{ECCV}, pages 360--377. Springer, 2024{\natexlab{c}}.

\bibitem[Yang et~al.(2024)Yang, Yang, Zhang, Hui, Zheng, Yu, Li, Liu, Huang,
  Wei, et~al.]{yang2024qwen2}
An Yang, Baosong Yang, Beichen Zhang, Binyuan Hui, Bo Zheng, Bowen Yu,
  Chengyuan Li, Dayiheng Liu, Fei Huang, Haoran Wei, et~al.
\newblock Qwen2. 5 technical report.
\newblock \emph{arXiv preprint arXiv:2412.15115}, 2024.

\bibitem[Ye et~al.()Ye, Xu, Liu, Hu, Yan, Qian, Zhang, Huang, and
  Zhou]{ye2024mplug}
Jiabo Ye, Haiyang Xu, Haowei Liu, Anwen Hu, Ming Yan, Qi Qian, Ji Zhang, Fei
  Huang, and Jingren Zhou.
\newblock mplug-owl3: Towards long image-sequence understanding in multi-modal
  large language models.
\newblock In \emph{ICLR}.

\bibitem[Ying et~al.(2021)Ying, Mandal, Ghadiyaram, and Bovik]{ying2021patch}
Zhenqiang Ying, Maniratnam Mandal, Deepti Ghadiyaram, and Alan Bovik.
\newblock Patch-vq:'patching up'the video quality problem.
\newblock In \emph{CVPR}, pages 14019--14029, 2021.

\bibitem[You et~al.(2024{\natexlab{a}})You, Gu, Li, Cai, Zhu, Dong, and
  Xue]{you2024descriptive}
Zhiyuan You, Jinjin Gu, Zheyuan Li, Xin Cai, Kaiwen Zhu, Chao Dong, and Tianfan
  Xue.
\newblock Descriptive image quality assessment in the wild.
\newblock \emph{arXiv preprint arXiv:2405.18842}, 2024{\natexlab{a}}.

\bibitem[You et~al.(2024{\natexlab{b}})You, Li, Gu, Yin, Xue, and
  Dong]{you2024depicting}
Zhiyuan You, Zheyuan Li, Jinjin Gu, Zhenfei Yin, Tianfan Xue, and Chao Dong.
\newblock Depicting beyond scores: Advancing image quality assessment through
  multi-modal language models.
\newblock In \emph{ECCV}, pages 259--276. Springer, 2024{\natexlab{b}}.

\bibitem[Zhai et~al.(2023)Zhai, Mustafa, Kolesnikov, and
  Beyer]{zhai2023sigmoid}
Xiaohua Zhai, Basil Mustafa, Alexander Kolesnikov, and Lucas Beyer.
\newblock Sigmoid loss for language image pre-training.
\newblock In \emph{ICCV}, pages 11975--11986, 2023.

\bibitem[Zhang et~al.(2024{\natexlab{a}})Zhang, Liu, Cherry, and
  Firat]{zhang2024scaling}
Biao Zhang, Zhongtao Liu, Colin Cherry, and Orhan Firat.
\newblock When scaling meets llm finetuning: The effect of data, model and
  finetuning method.
\newblock In \emph{ICLR}, 2024{\natexlab{a}}.

\bibitem[Zhang et~al.(2024{\natexlab{b}})Zhang, Wu, Li, Li, Ma, Liu, and
  Li]{zhang2024video}
Yuanhan Zhang, Jinming Wu, Wei Li, Bo Li, Zejun Ma, Ziwei Liu, and Chunyuan Li.
\newblock Video instruction tuning with synthetic data.
\newblock \emph{CoRR}, 2024{\natexlab{b}}.

\bibitem[Zhang et~al.(2025)Zhang, Jia, Wu, Li, Chen, Zhou, Sun, Liu, Min, Lin,
  et~al.]{zhang2024q}
Zicheng Zhang, Ziheng Jia, Haoning Wu, Chunyi Li, Zijian Chen, Yingjie Zhou,
  Wei Sun, Xiaohong Liu, Xiongkuo Min, Weisi Lin, et~al.
\newblock Q-bench-video: Benchmarking the video quality understanding of lmms.
\newblock \emph{CVPR}, 2025.

\bibitem[Zhou et~al.(2023)Zhou, Liu, Xu, Iyer, Sun, Mao, Ma, Efrat, Yu, Yu,
  et~al.]{zhou2023lima}
Chunting Zhou, Pengfei Liu, Puxin Xu, Srinivasan Iyer, Jiao Sun, Yuning Mao,
  Xuezhe Ma, Avia Efrat, Ping Yu, Lili Yu, et~al.
\newblock Lima: Less is more for alignment.
\newblock \emph{NIPS}, 36:\penalty0 55006--55021, 2023.

\bibitem[Zhou et~al.(2024)Zhou, Wang, Lin, Su, Chen, Tao, Zheng, Yuan, Wan, and
  Zhang]{zhou2024uniaa}
Zhaokun Zhou, Qiulin Wang, Bin Lin, Yiwei Su, Rui Chen, Xin Tao, Amin Zheng, Li
  Yuan, Pengfei Wan, and Di Zhang.
\newblock Uniaa: A unified multi-modal image aesthetic assessment baseline and
  benchmark.
\newblock \emph{arXiv preprint arXiv:2404.09619}, 2024.

\end{thebibliography}
}
\maketitlesupplementary
\appendix
\section{Experiments Supplementary Materials}
\subsection{System prompts for training and evaluation}
\label{system prompts}
In all training process and task evaluations, we set unified system prompts (prefix) for all LMM models, which is: \textit{``You will receive a keyframe sequence sampled at an average of one frame per second from a video of {length} seconds, with the keyframe sequence ordered in alignment with the video’s temporal order. Please answer the following question based on the information provided."} For our model, since motion feature extraction is involved, the system prompt also includes: \textit{``In addition, you will receive a motion feature sequence that corresponds to the number of frames in the video, {num of frames}."} For the spatiotemporal fine-grained tasks, which involve specific time-related questions, we introduce a standardized representation of time to ensure consistency in the responses from different models. The time point representation rule is as follows: \textit{``When the video starts playing, this timepoint is denoted as ``1 second". When the 1st second ends and the 2nd second begins, this timepoint is marked as ``2 seconds', and so on."}.

\subsection{Evaluation details}
\label{evaluation}
\paragraph{Quality rating evaluation details.}
We adopt the following procedure to assess the quality during evaluation:

\[
\mathcal{Q} = \sum_{i=1}^5 \omega_i \frac{e^{\mathcal{P}_{\textit{quality\_levels}[i]}}}{\sum_{i=1}^5 e^{\mathcal{P}_{\textit{quality\_levels}[i]}}},
\]
where \textit{quality\_levels} refers to a list of predefined quality categories: \textit{[Excellent, Good, Fair, Poor, Low]}, and \(\mathcal{P}\) denotes the model's logit outputs for each respective quality level. Specifically, the vector corresponding to the quality description word in the model’s output sequence is first extracted, where its dimension matches the tokenizer's vocabulary size (located at the $-3$ index in our model). The logit values at the specific indices of this vector, which correspond to the $5$ quality level in the tokenizer’s vocabulary (indices $1550$, $1661$, $6624$, $7852$, and $3347$ in our model), are then selected. These logits are subsequently normalized using the softmax function. 

The values \(\omega\) represent the weight factors assigned to the normalized probabilities of each quality level, given by $[1, 0.75, 0.5, 0.25, 0]$. The resulting weighted sum of these probabilities produces the predicted quality score \(\mathcal{Q}\), which is confined within the range of $[0,1]$.

We directly call \textit{model.forward()} with the prompt extended to the word just before outputting the quality level, i.e., the input prompt would be:

\textit{How would you rate the quality of this video? The quality of this video is:"}

The final output corresponds to the token after the last newline or special token (like EOS) in the original prompt, which represents the quality level.

\paragraph{Quality understanding evaluation details.}
For the quality understanding task, we use \textit{model.generate()} with \textit{greedy search} to ensure the reproducibility of the results. For multiple-choice questions in the benchmarks, we compare the first letter of the output (usually the option selected) with the correct answer and output the accuracy. For open-ended questions and multiple-choice questions where the first letter is not an option, we use \textit{sota LMM} for judgment. For multiple-choice questions, we directly assess whether the answer is correct (scoring 0 or 1). For open-ended questions, we evaluate them based on $3$ criteria: \textit{completeness}, \textit{accuracy}, and \textit{relevance}, with a score given as 0, 1, or 2. The specific evaluation standards are as follows:

\textit{``Given the {question}, evaluate whether the response {answer} completely matches the correct answer {correct ans}. 
First, check the response and please rate score 0 if the response is not a valid answer.
Please rate score 2 if the response completely or almost completely matches the correct answer on completeness, accuracy, and relevance. 
Please rate score 1 if the response partly matches the correct answer on completeness, accuracy, and relevance.
Please rate score 0 if the response doesn't match the correct answer on completeness, accuracy, and relevance at all.
Please provide the result in the following format: Score:"}

We set up $5$ rounds of \textit{sota LMM} scoring for each question. The final score for the question is determined by ``majority voting", selecting the most frequently occurring score. Based on our experiments, there has been no instance where the score distribution resulted in a ``2/2/1" split.

\subsection{Model structure/Training hyper-parameters}
The model structure and training hyper-parameters are detailed in Tab. \ref{tab:hyperparam}. 
\begin{table*}[h]
 \renewcommand\arraystretch{1.3}
\renewcommand\tabcolsep{6pt}
\belowrulesep=0pt\aboverulesep=0pt
\centering
\caption{
Details of the model structure and hyper-parameters for the model training. The \textcolor{red}{red} /  \textcolor{blue}{blue} colors represent the different hyperparameters used in the first and second rounds of complementary training, respectively. Other entries without color differentiation indicate that the hyperparameter remains consistent across both rounds of complementary training.
}
\resizebox{\linewidth}{!}{
\begin{tabular}{l| c| c}
\toprule
\textbf{Model Structure/Training Hyper-Parameters} &  \textbf{Name/Value} &  \textbf{More Information}  \\
\midrule
Vision Tower & \textit{SigLIP-SO400m} &\textit{Parameter size=}$397.75$M, \textit{Tokens per keyframe=$196$}  \\
Vision Projector & \textit{2-layers MLP+GeLU}&\textit{Parameter size}=$16.98$M \\
Motion Extractor& \textit{SlowFast-R50} &\textit{Parameter size=}$34.16$M, only use the fast path feature  \\
Motion Projector& \textit{2-layers MLP+GeLU}&\textit{Parameter size}=$13.77$M, the same with the structure of vision projector\\
LLM init. & \textit{Qwen-2 (7B)} &Decoder-only model, \textit{parameter size}=$7660.56$M \\
Keyframes Sampling Interval &1 second&/\\   
Keyframes Resolution         & $336\times 336$&/\\

Frames (for motion extraction) Resolution         & $224\times 224$&/\\
Batch Size (videos) & 8&\textit{Per device train batch size=1} \\
LR Max & \textcolor{red}{1e-5} / \textcolor{blue}{1e-6} &/ \\
LR Schedule & cosine decay&/ \\
Warmup Epochs & $0.03$&/ \\
Weight Decay & $0$&/ \\
Gradient Accumulation Steps   & $\textcolor{red}{1}$ / $\textcolor{blue}{2}$&/ \\
Numerical Precision      & $\mathtt{bfloat16}$&/ \\
Epoch &\textcolor{red}{$1$} / \textcolor{blue}{$2$}& / \\
Optimizer & AdamW&/ \\
Activation Checkpointing &\checkmark&/ \\
Deepspeed Stage & $3$ &/ \\
\bottomrule
\end{tabular}
}

\label{tab:hyperparam}

\end{table*}
\subsection{Additional analysis for experiments results}
\label{analysis}

\begin{table*}[t]\tiny
    \centering
    
    \renewcommand\arraystretch{1.2}
    \renewcommand\tabcolsep{6pt}
    \belowrulesep=0pt\aboverulesep=0pt

    \caption{Performance of different training data for quality rating. }
        \vspace{-5pt}
    \resizebox{1\linewidth}{!}
    {\begin{tabular}{l|cc|cc|cc|cc|cc|cc}
 \hline
    \multicolumn{1}{l|}{\textbf{Training Datasets}}&\multicolumn{2}{c|}{\textbf{LSVQ(1080p)}}& 
    \multicolumn{2}{c|}{\textbf{LSVQ(test)}}&\multicolumn{2}{c|}{\textbf{LIVE-VQC}}&\multicolumn{2}{c|}{\textbf{KoNViD-1k}}&\multicolumn{2}{c|}{\textbf{YT-UGC}}&\multicolumn{2}{c}{\textbf{MOS-20K}}\\ 
    \cdashline{1-13}
       \multicolumn{1}{l|}{\textit{\textbf{Metrics}}}&SRCC &PLCC &SRCC &PLCC &SRCC &PLCC &SRCC &PLCC &SRCC &PLCC&SRCC &PLCC \\ \cdashline{1-13}
    
      \textit{LSVQ(train)}&\textcolor{red}{0.829}&\textcolor{red}{0.849}&\textcolor{red}{0.904} &0.893&\textcolor{red}{0.857}&\textcolor{red}{0.872}&0.892&0.886 &0.826&0.835 &0.802&0.791 \\
      \textit{Merged} &0.815&0.838&0.902&\textcolor{red}{0.905} &0.826&0.855&\textcolor{red}{0.895}&\textcolor{red}{0.900} &\textcolor{red}{0.872}&\textcolor{red}{0.873}&\textcolor{red}{0.837}&\textcolor{red}{0.837} \\
     
  \hline
    \end{tabular}}
      \vspace{-8pt}
    \label{tab:merged_rating}
\end{table*}
\paragraph{Effects of merged training data for quality rating.}
We compared the \textbf{rater} trained on the  merged dataset (the combination of LSVQ(train) and OmniVQA-MOS-20K(train))  and LSVQ (train), with evaluation results on $6$ test datasets. The results are presented in Tab. \ref{tab:merged_rating}.

The effectiveness of the merged data varies depending on the evaluation dataset. We assume that the performance improvement from combined training is related to the similarity between the test set and the video content of OmniVQA-MOS-20K(train). Most notably, on the YT-UGC dataset, the merged training yields the most significant performance improvement. This is clearly due to the high consistency between the video sources of the YT-UGC dataset and OmniVQA-MOS-20K(train) (both are entirely or mostly sourced from UGC videos on the \textit{YouTube} platform). This consistency leads to a strong alignment of video quality priors between OmniVQA-MOS-20K(train) and YT-UGC, thus producing the best performance when trained with the merged data. For the KoNViD-1k and LSVQ(test), since the results obtained from LSVQ(train) training are already quite excellent (we believe close to the highest performance achievable for these datasets), the data augmentation does not lead to significant performance improvements.

\paragraph{Is model-size scaling law evident in the video quality understanding tasks?}
For open-source LMMs designed for general visual understanding tasks, the model-size scaling law is not evident in quality understanding tasks. This is observed in \textit{Internvl-2}, \textit{LLaVa-Onevision}, and \textit{Qwen2.5-vl}. These models exhibit a clear parameter scaling law in general visual question answering tasks (40B or 72B models significantly outperform the 7B or 8B models). However, in video quality understanding tasks, whether general or fine-grained, increasing the model size does not result in significant performance improvements. The only exception is \textit{Qwen2-vl}, where the \textit{Qwen2-vl (72B)} outperforms the \textit{Qwen2-vl (7B)} in both general and fine-grained tasks, making it the best-performing open-source LMM. Based on this observation, we hypothesize that in downstream tasks like quality understanding, the scaling law of model parameters is closely related to the amount of data directly associated with or similar to the task during training. Different training data or strategies can affect the scaling effect of model parameters in such downstream tasks.

\paragraph{Deeper analysis on training strategies.}
We find that the difference between direct training and complementary training is more pronounced in the quality rating task, while the difference between mixed training and complementary training is more prominent in the quality understanding tasks. The former is easier to explain, as the amount of training data for the quality rating task is much smaller than that for the quality understanding task. Therefore, the pretraining effect for the quality understanding data is evidently stronger, leading to amore noticeable differences in the quality rating task. For the latter, we hypothesize that the task objective in the quality rating task is much more direct compared to that in the quality understanding task, and its data composition is much simpler. Consequently, training with these relatively simple and consistent data structures, as opposed to the complex and varied data for the quality understanding task, reduces the model's ability to generalize across various types of quality understanding tasks and hinders the diversity of its output.

For example, in the \textbf{OmniVQA-MOS-20K} dataset, the Q\&A pairs have questions like \textit{``How would you rate the quality of this video?"} with answers corresponding to the quality level of the video. In contrast, in the \textbf{OmniVQA-Chat-400K} technical branch, many Q\&A pairs contain questions like \textit{``How would you evaluate the quality of this video?"}, where the answers are the 
summarized video-level quality descriptions. While these two questions are semantically similar, their answers emphasize different aspects in their respective datasets. 

Clearly, during the \textbf{mix} training, where both types of questions are included in the same batch, the training strategy could negatively impact the model’s learning direction. However, under the \textbf{complementary} training, these two types of questions are not mixed within the same batch, effectively preventing mutual interference.

An alternative approach is to prepend a \textbf{task-type descriptor} (e.g., "rating" or "understanding") to differentiate between these tasks and guide the model’s learning. However, this method would also require adding such prefixes during inference, increasing the computational cost.
\vspace{-8pt}
\section{Additional Statistical Information}
\subsection{VQA datasets summary}
Tab. \ref{tab:VQA_summary} presents the statistical information of mainstream (UGC) VQA video datasets, sorted in chronological order of publication. This serves as supporting evidence for our assertions in Sec. \ref{sec:intro}. As observed, the majority of these VQA datasets lack the instruction component, making them unsuitable as direct training data for video quality understanding models. Additionally, the number of videos in these datasets rarely exceeds $20$K. In contrast, the OmniVQA dataset series — particularly the OmniVQA-Chat-400K — stands out as the  \textbf{largest} in scale to date in terms of the \textbf{number of videos}, the \textbf{variety of covered tasks}, and the \textbf{total number of instruction pairs}.

\subsection{Statistical information for each branch}
\begin{figure*}[t]
    \centering
    \vspace*{-0.8cm}
    \includegraphics[width=\linewidth, page=1]{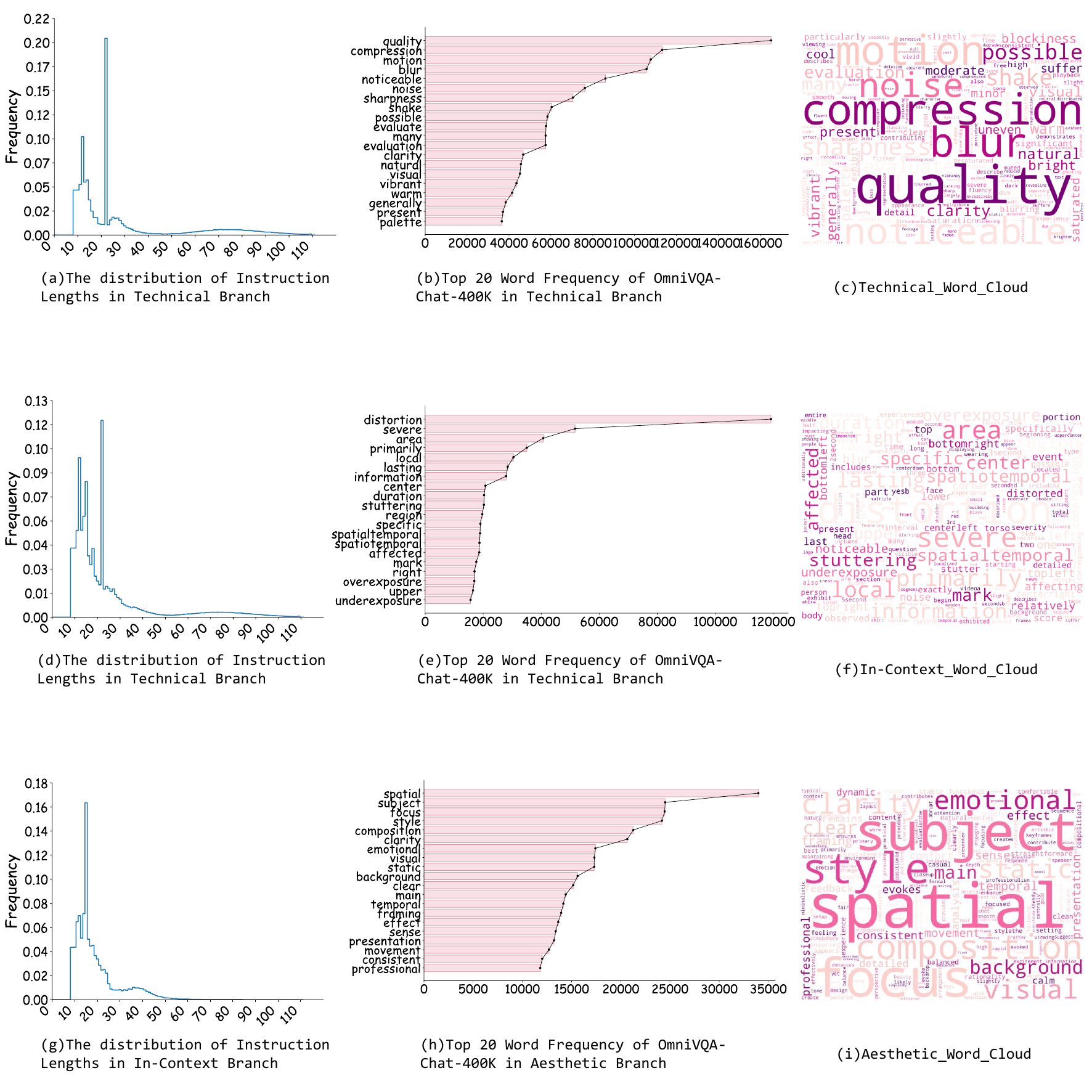}
    \caption{Additional Statistical Information}
    \label{fig:branch-material-p0}
\end{figure*}

Fig. \ref{fig:branch-material-p0} displays the distribution of instruction lengths, the top 20 word frequencies, and the word cloud of the $3$ branches in \textbf{OmniVQA-Chat-400K}.
\begin{table*}[h]
    \centering
    \caption{Summary of various VQA datasets.}
    \vspace{-7pt}
    \small
    \renewcommand\arraystretch{1.2}
    \renewcommand\tabcolsep{1pt}
    \belowrulesep=0pt\aboverulesep=0pt
    \begin{tabular}{l|c|c|c|c|c|c}
        \toprule
        \textit{Datasets for VQA} &\# Videos & MOS &\# MOS &Instruction &\# Instruction& Description \\ 
        \midrule
        LIVE-VQA \cite{} & 160 & \ding{51} &160 & \ding{55} & / &Full-reference video quality rating  \\ 
        CVD2014&234 & \ding{51} &234 & \ding{55} & / & Quality assessment of video captured by cameras \\ 
        LIVE-Qualcomm&208 & \ding{51} &208 & \ding{55} & / & Mobile in-capture video quality rating \\ 
        KoNViD-1K& 1,200 & \ding{51} &1,200 & \ding{55} & /  & Unified UGC video quality rating \\ 
        LIVE-VQC& 585 & \ding{51} & 585 & \ding{55} & / & Quality rating of real world UGC videos \\ 
        YouTube-UGC& 1,380 & \ding{51} & 1,380  & \ding{55} & /& Quality rating of UGC videos \\ 
        LSVQ&39,075 & \ding{51} &39,075  & \ding{55} & / &Large-scale quality rating of UGC videos \\ 
        LIVE-NFLX-I&558 & \ding{51} & 558  & \ding{55} & / & Quality-of-experience (QoE) rating of hand-craft streaming videos \\ 
        LIVE-NFLX-II&420 & \ding{51} & 420  & \ding{55} & / & QoE rating of real-world streaming videos\\ 
        WaterlooSQoE-III& 450 & \ding{51} & 450  & \ding{55} & / & QoE rating of hand-craft streaming videos\\ 
        LBVD& 1,013 & \ding{51} & 1,013  & \ding{55} & / & QoE assessment of in-the-wild streaming videos \\ 
        WaterlooSQoE-IV& 1,350 & \ding{51} & 1,350  & \ding{55} & / & Large-scale QoE assessment of hand-craft streaming videos \\ 
        TaoLive& 3,762 & \ding{51} &3,762  & \ding{55} & / &Quality rating of live streaming (compresqsed) videos \\ 
        Maxwell& 4,543 & \ding{51} &4,543  & \ding{55} & / &Fine-grained (technical/aesthetic) quality rating of UGC videos \\ 
        VQA\textsuperscript{2}-Stage-1& 12,385 & \ding{55} & /  & \ding{51} &12,385 &Pre-training MIDB for distortion recognition \\ 
        VQA\textsuperscript{2}-Stage-2& 30,156 & \ding{51} & 30,156  & \ding{51} &30,156 &Large-scale MIDB specially for video quality rating. \\
        VQA\textsuperscript{2}-Stage-3& 15,500 & \ding{55} & /  & \ding{51} &115,214 &Human-annotated MIDB for video quality understanding. \\
        \textbf{OmniVQA-Chat-20K}& 20,000 & \ding{51} &20,000  & \ding{51} &20,000 &Large-scale MIDB for quality rating for in-the-wild UGC videos \\ 
        \textbf{OmniVQA-MOS-400K}& 86,716 & \ding{55} & /  & \ding{51} & 402,987 &Machine-dominated MIDB for video quality understanding. \\ 
        \bottomrule
    \end{tabular}
    \vspace{-6pt}
    \label{tab:VQA_summary}
\end{table*}

\section{Distorted Video Generation}

\label{Distorted}
\begin{figure*}[t]
    \centering
    \vspace*{-1cm}
    \includegraphics[width=\linewidth, page=1]{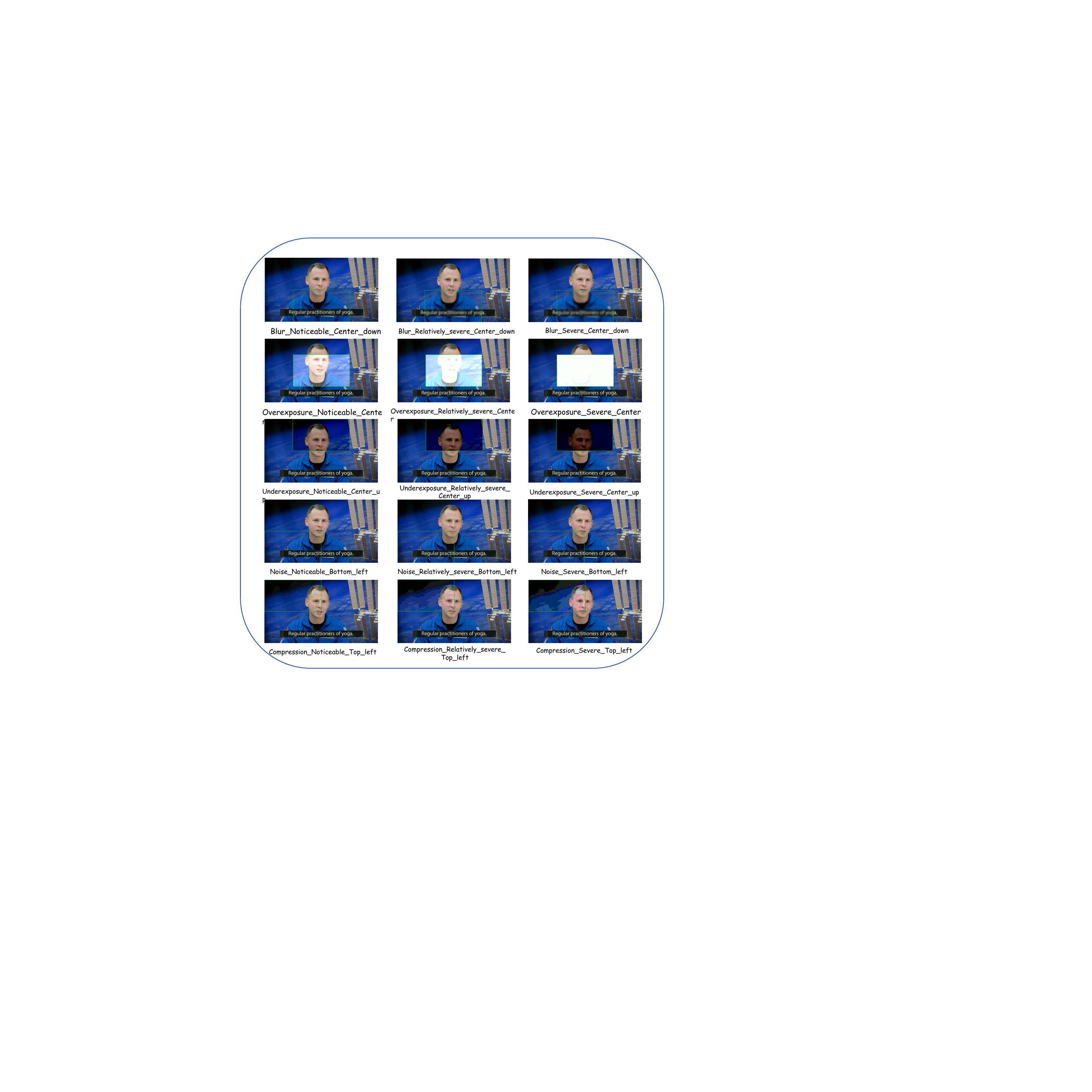}
    \caption{Spatial-distorted videos examples.}
    \label{fig:incontext_examples}
\end{figure*}

The method for constructing spatial distorted videos in the in-context branch (Sec. \ref{in-context}) is as follows, where ``level" refers to:  ``noticeable" (1),  ``relatively severe" (2), and  ``severe" (3).  The distorted region is selected in one of the following locations: \textbf{top-left}, \textbf{bottom-left}, \textbf{top-right}, \textbf{bottom-right}, \textbf{center}, \textbf{center-up}, \textbf{center-down}, \textbf{center-left}, \textbf{center-right}. The detailed examples of distortion types and location are illustrated in Fig. \ref{fig:incontext_examples}. \\
1. \textbf{Gaussian Blur (Blur)}: \\
The Gaussian blur effect smooths the image by applying a filter with a kernel size determined by the level. The larger the level, the stronger the blur effect.
$$
\footnotesize
\text{Blurred} = \text{GaussianBlur}(\text{Region}, (16 \times \text{Level} + 1, 16 \times \text{Level} + 1), 0)
$$
2. \textbf{Overexposure (Brightness Increase)}: 
Overexposure increases the brightness of each pixel by a factor of \(80 \times \text{level}\). The resulting pixel values are clamped to stay within the valid range of 1 to 254.
\[
\footnotesize
\text{overexposed}(x, y) = \min\left(\max\left(\text{region}(x, y) + 80 \times \text{level}, 1\right), 254\right)
\]
3. \textbf{Underexposure (Brightness Decrease)}: \\
Underexposure reduces the brightness of each pixel by a factor of \(40 \times \text{level}\). Similar to overexposure, the pixel values are clamped to ensure they remain within the range 1 to 254.
\[
\footnotesize
\text{underexposed}(x, y) = \min\left(\max\left(\text{region}(x, y) - 40 \times \text{level}, 1\right), 254\right)
\]
4. \textbf{Noise (Gaussian Noise)}: \\e
The noise distortion adds Gaussian noise to the image, where the standard deviation of the noise is determined by \( \sigma = \sqrt{250 \times \text{level}} \). This introduces random variations in pixel values, creating a noisy effect.
\[
\text{noise}(x, y, c) = \text{region}(x, y, c) + \mathcal{N}(0, \sigma^2)
\]
where \( \mathcal{N}(0, \sigma^2) \) represents a Gaussian distribution with mean 0 and variance \( \sigma^2 \). \\
5. \textbf{Compression Distortion (JPEG Compression)}: \\
The compression distortion simulates the loss of quality due to JPEG compression. The quality of the compressed image is controlled by the level, where higher levels result in lower quality (lower JPEG quality). The compression quality is calculated as \( \max(15 - 5 \times \text{level}, 1) \).
\[
\footnotesize
\text{compressed}(x, y) = \text{JPEG}(\text{region}, \text{quality} = \max(15 - 5 \times \text{level}, 1))
\]
\section{Machine Annotation Process Detail}
\label{process}
\begin{figure*}[!htbp]
    \centering
    \vspace*{-0.8cm}
    \includegraphics[width=0.8\linewidth, page=1]{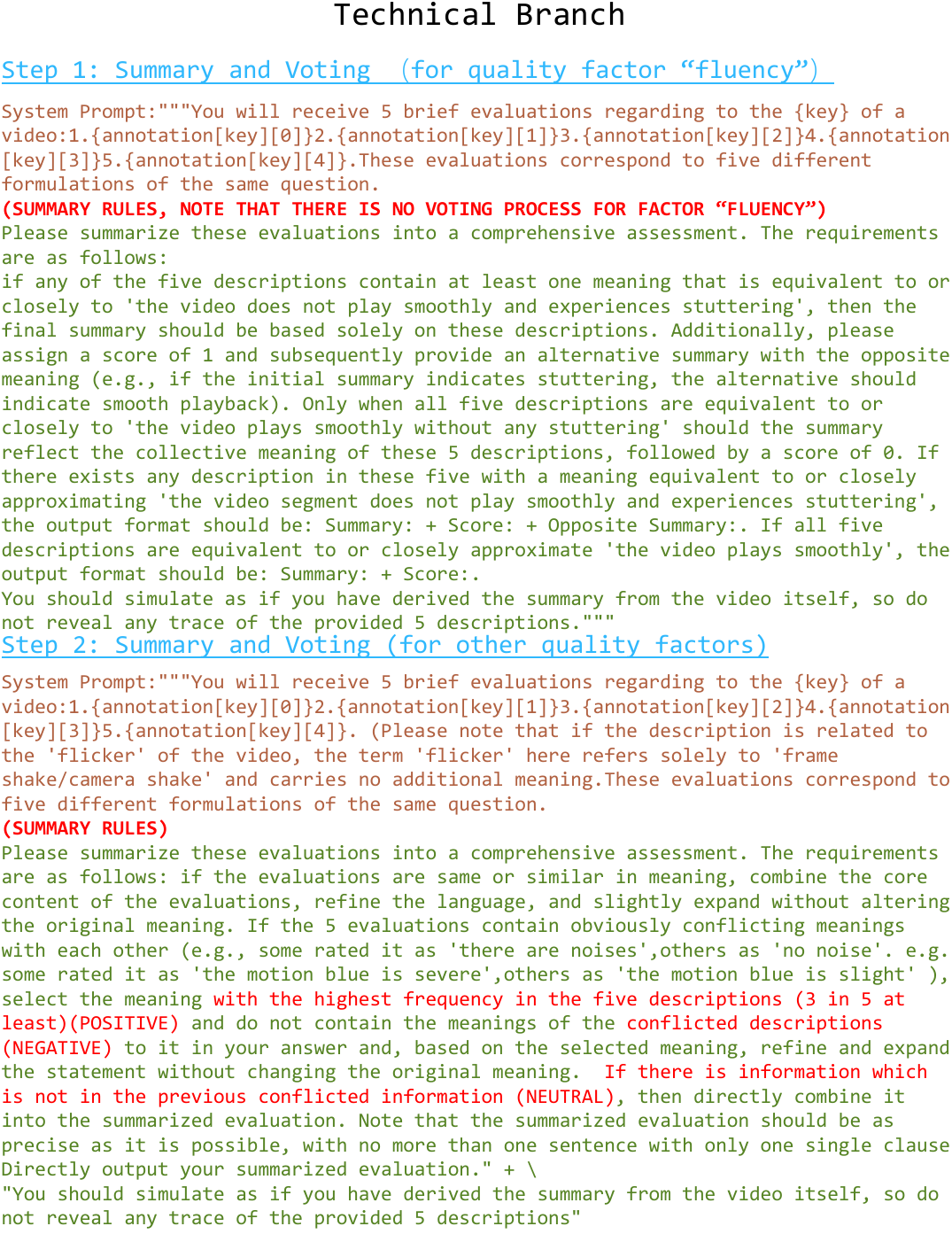}  
    \caption{Annotation prompts for technical branch (1). \\The highlighted content in \textcolor{red}{red} represents the summary or key points of each step.
}
    \label{fig:branch-material-p1}
\end{figure*}

\begin{figure*}[!htbp]
    \centering
    \vspace*{-0.8cm}
    \includegraphics[width=0.8\linewidth, page=2]{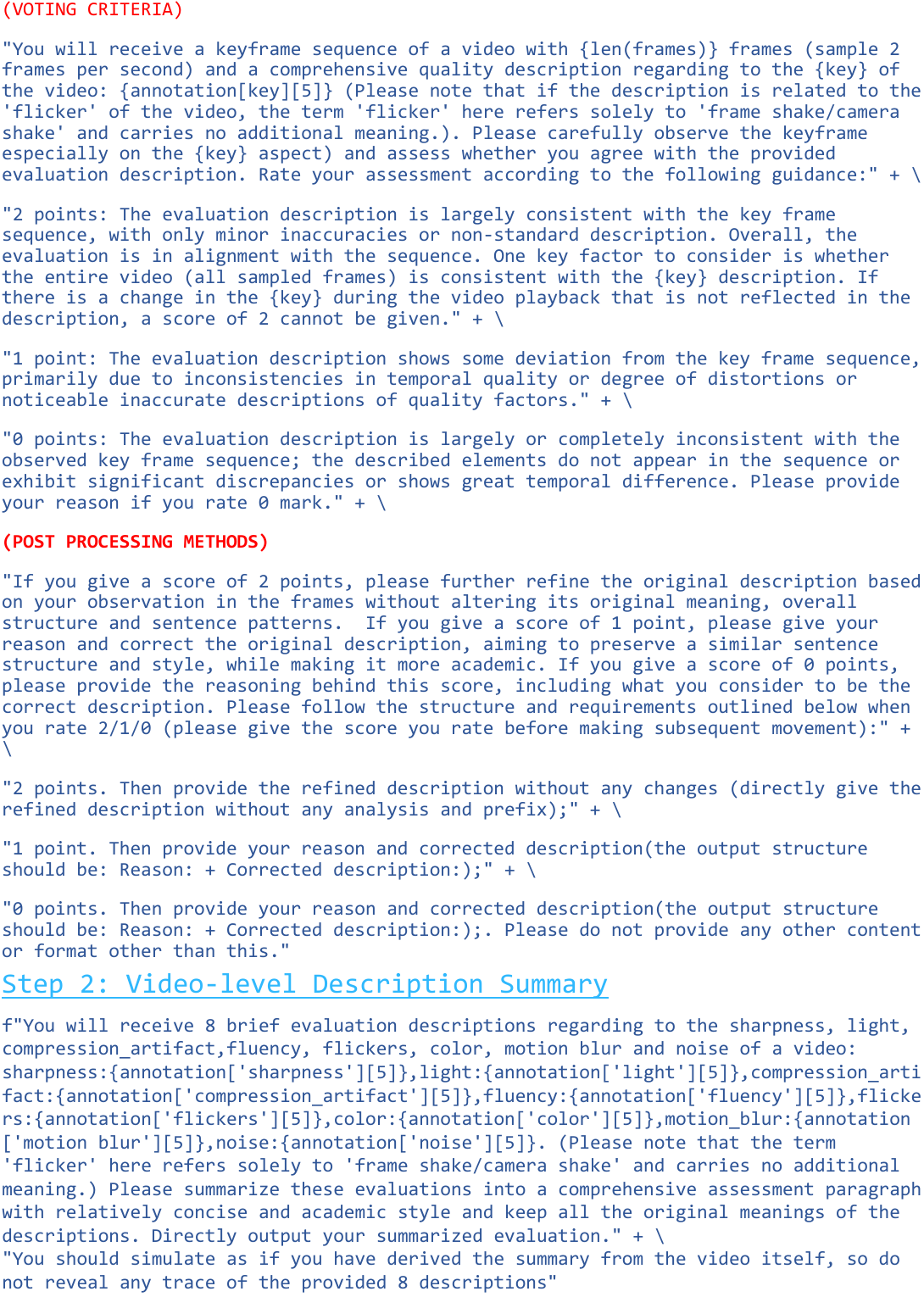}
    \caption{Annotation prompts for technical branch (2).}
    \label{fig:branch-material-p2}
\end{figure*}

\begin{figure*}[!htbp]
    \centering
    \vspace*{-0.8cm}
    \includegraphics[width=0.8\linewidth, page=3]{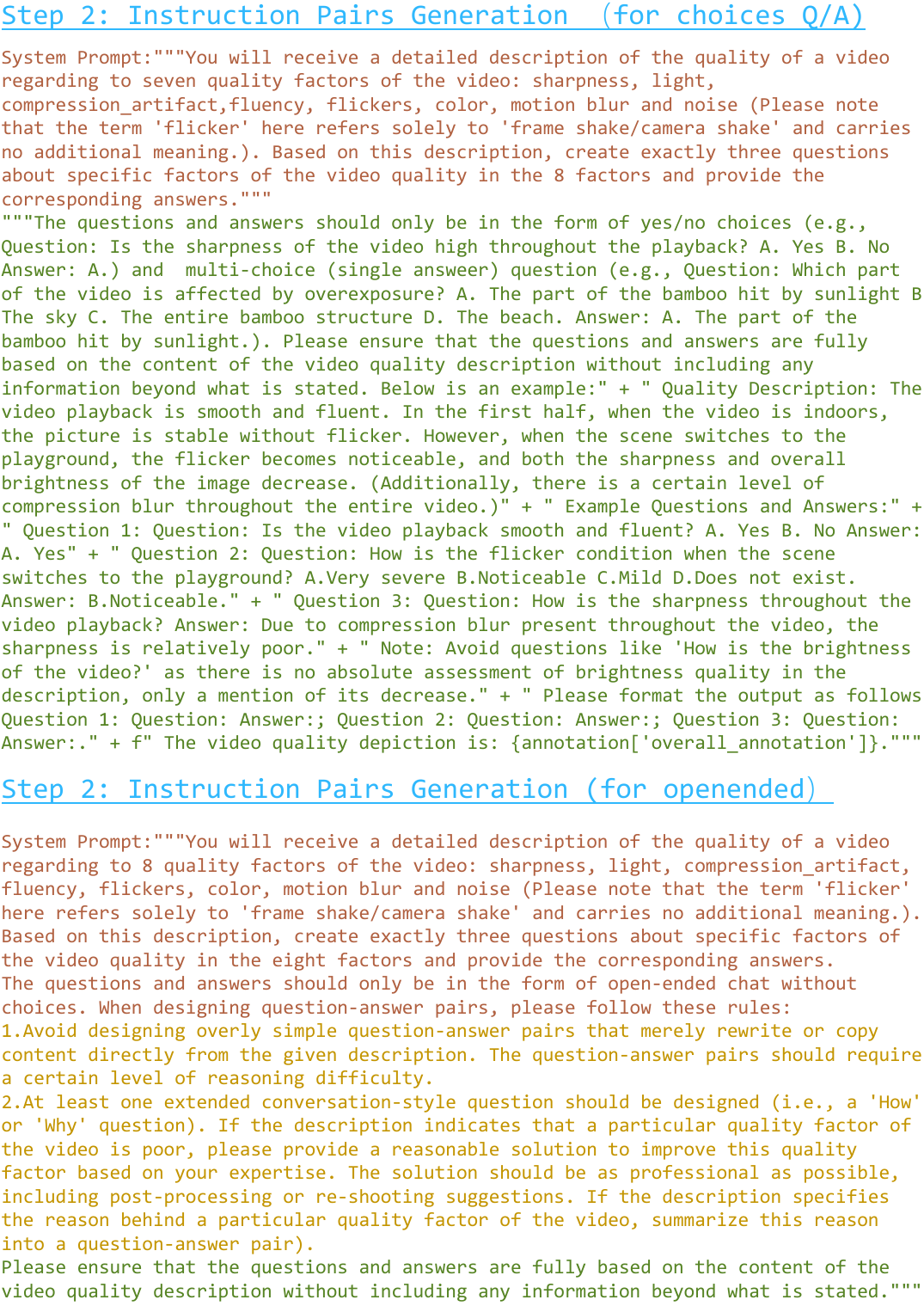}
    \caption{Annotation prompts for technical branch (3).}
    \label{fig:branch-material-p3}
\end{figure*}

\begin{figure*}[!htbp]
    \centering
    \vspace*{-0.8cm}
    \includegraphics[width=0.78\linewidth, page=4]{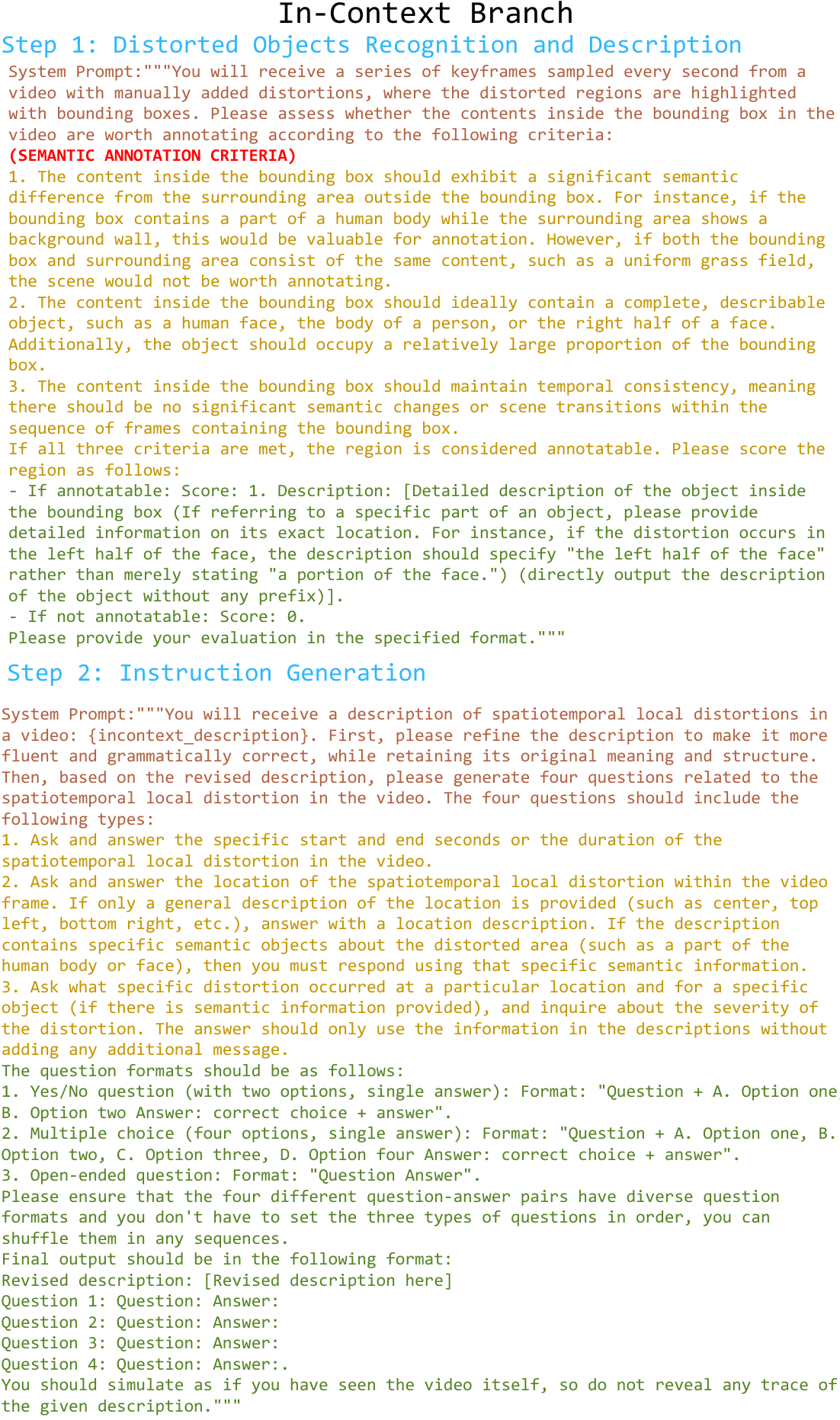}
    \caption{Annotation prompts for in-context branch}
    \label{fig:branch-material-p4}
\end{figure*}

\begin{figure*}[!htbp]
    \centering
    \vspace*{-0.8cm}
    \includegraphics[width=0.8\linewidth, page=5]{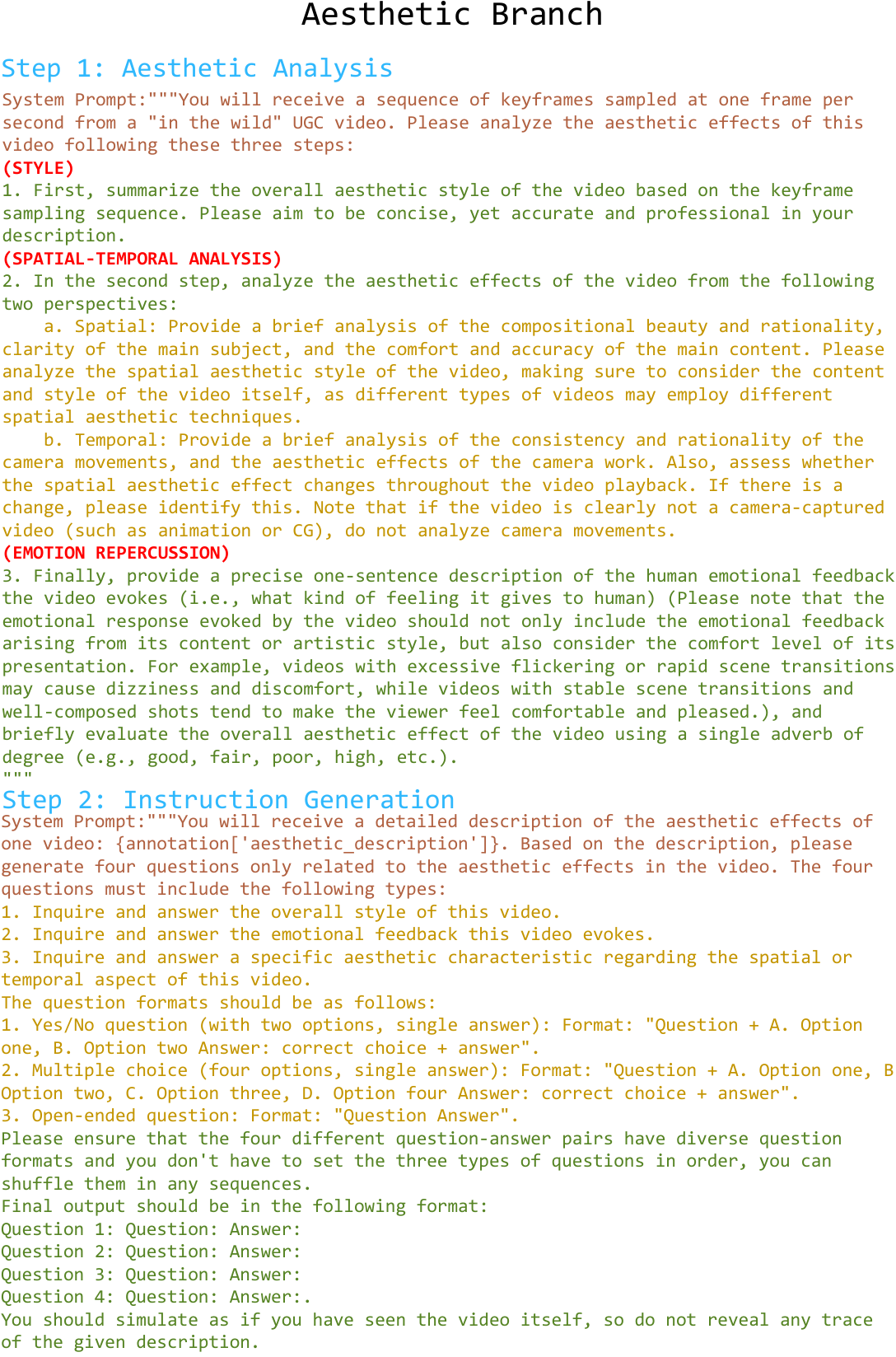}
    \caption{Annotation prompts for aesthetic branch.}
    \label{fig:branch-material-p5}
\end{figure*}

\paragraph{Annotation prompts for each branch}
Figs. \ref{fig:branch-material-p1},\ref{fig:branch-material-p2}, and \ref{fig:branch-material-p3} illustrate the technical branch prompts and detailed annotation pipeline. 
Fig. \ref{fig:branch-material-p4} illustrates the in-context branch annotation prompts.
Fig. \ref{fig:branch-material-p5} illustrates the aesthetic branch annotation prompts.

\begin{figure*}[!htbp]
    \centering
    \vspace*{-0.8cm}
    \includegraphics[width=0.78\linewidth, page=1]{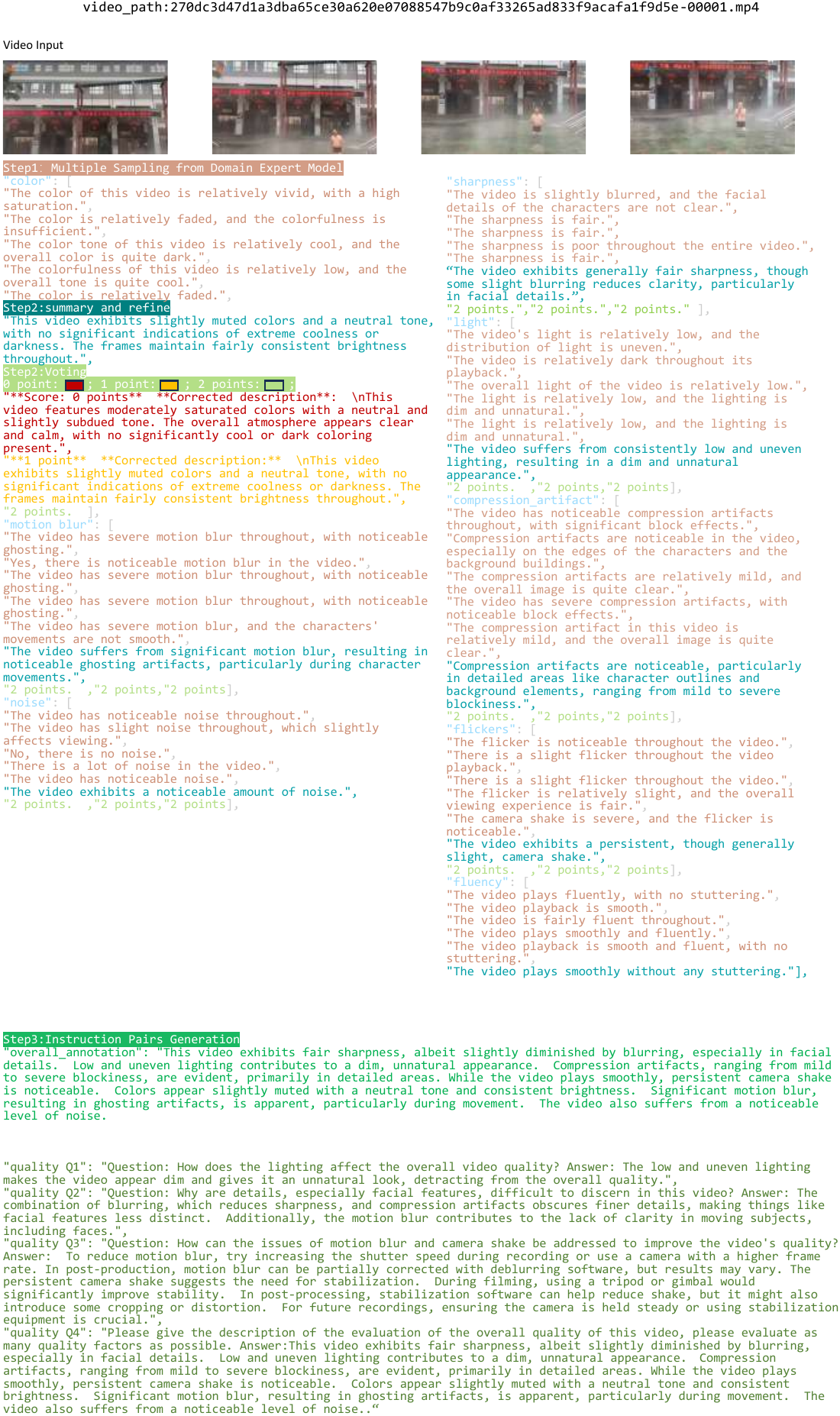 }
    \caption{Technical branch annotation example(1).  (Note that the ``summary and refine” process is done after the human-in-the-loop selection process if there is ``0" in voting results.)}
    \label{fig:technical-branch-p1}
\end{figure*}

\begin{figure*}[!htbp]
    \centering
    \vspace*{-0.8cm}
    \includegraphics[width=0.8\linewidth, page=2]{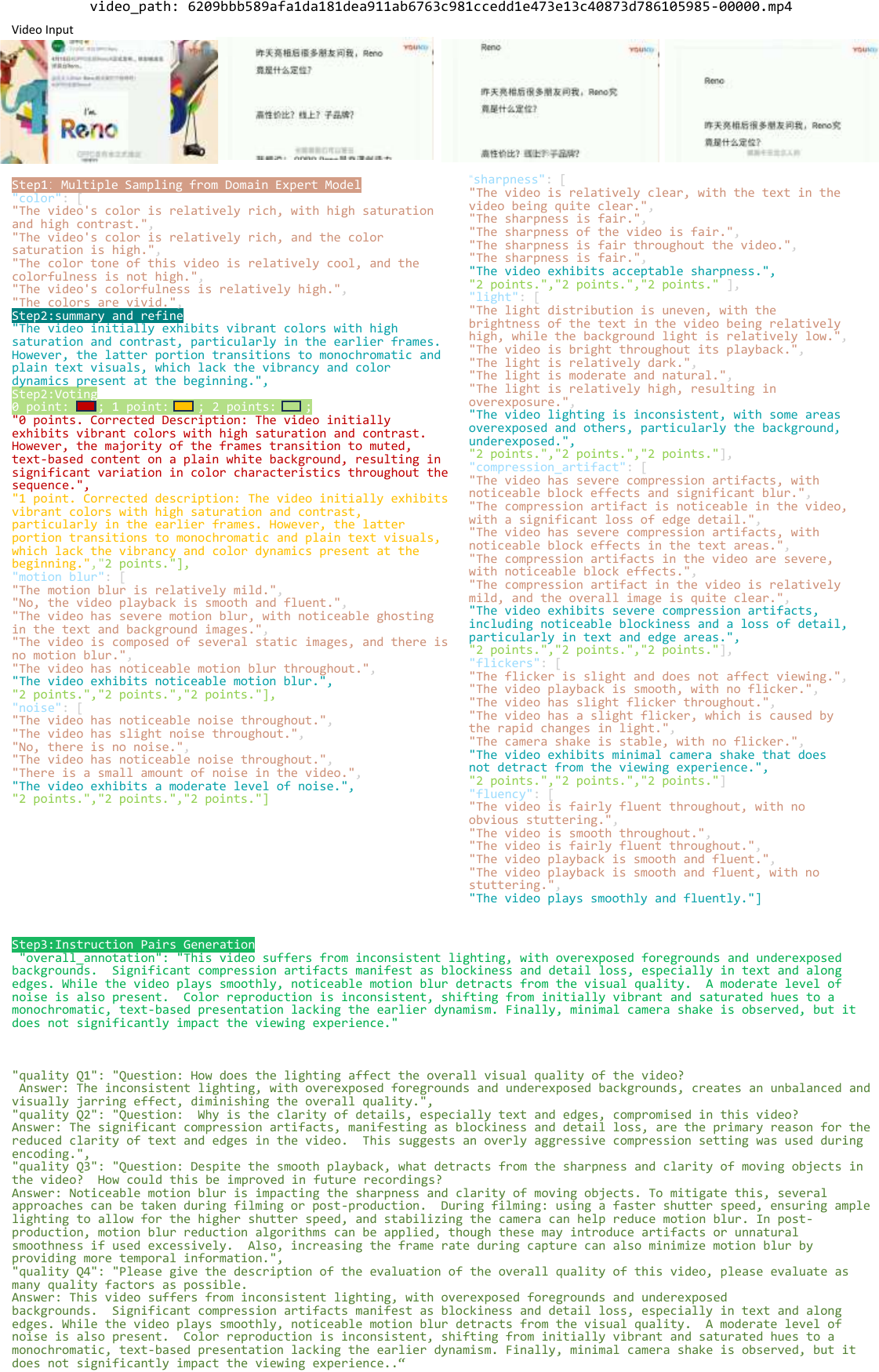}
    \caption{Technical-branch annotation example(2). (Note that the ``summary and refine” process is done after the human-in-the-loop selection process if there is ``0" in voting results.)}
    \label{fig:technical-branch-p2}
\end{figure*}

\begin{figure*}[!htbp]
    \centering
    \vspace*{-0.8cm}
    \includegraphics[width=0.75\linewidth, page=3]{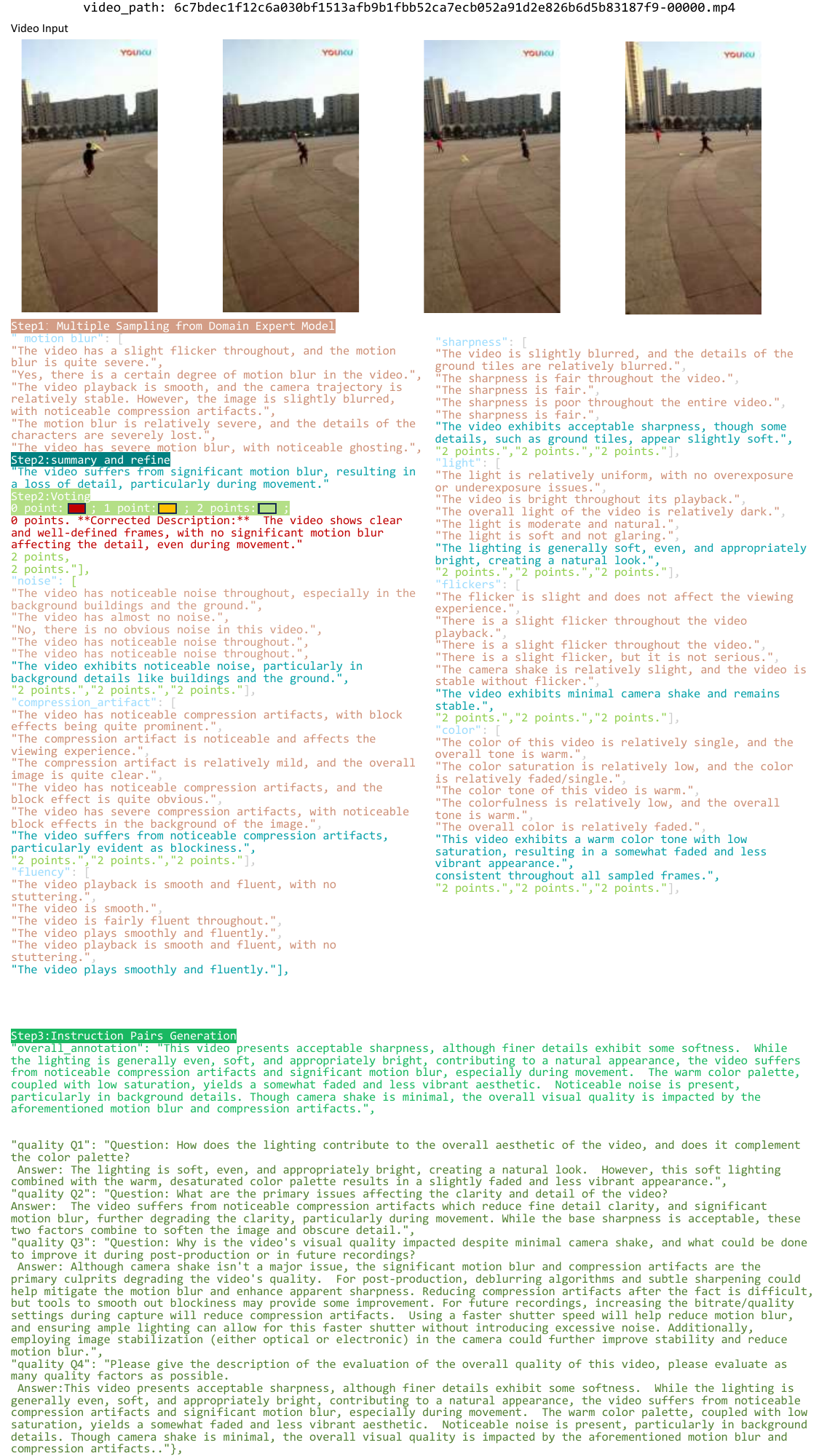}
    \caption{Technical branch annotation example(3).}
    \label{fig:technical-branch-p3}
\end{figure*}

\paragraph{Technical branch examples}
Figs. \ref{fig:technical-branch-p1}, \ref{fig:technical-branch-p2}, and \ref{fig:technical-branch-p3} show $3$ annotated examples of the technical branch.

\begin{figure*}[t]
    \centering
    \includegraphics[width=0.8\linewidth]{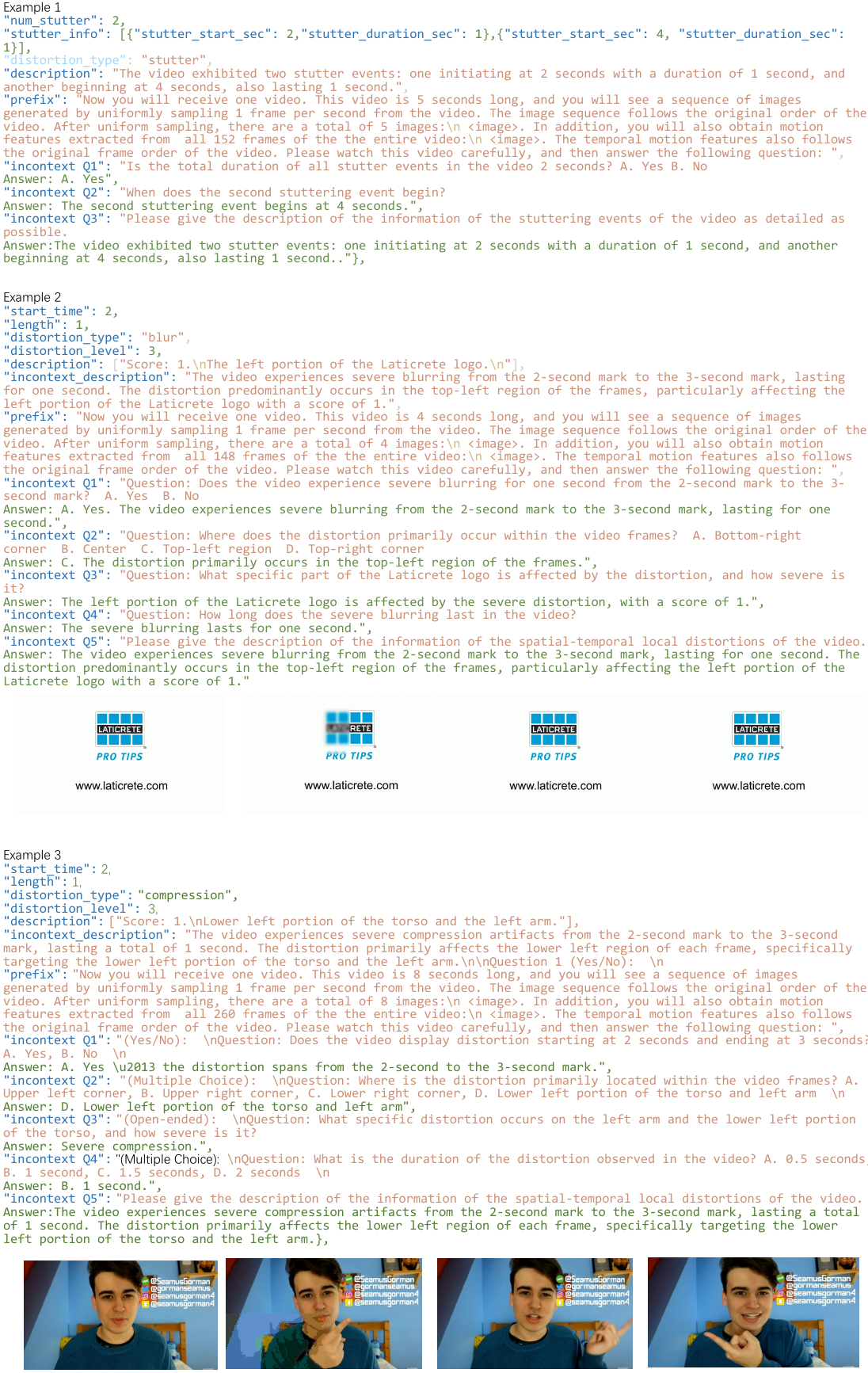}
    \caption{In-context branch annotation examples. Since frame freezing effects cannot be demonstrated through screenshots, we provide only an annotation example for the stutter distortion.
}
    \label{fig:incontext-branch}
\end{figure*}

\paragraph{Incontext branch examples}
Fig.~\ref{fig:incontext-branch} illustrates $3$ annotated examples from the in-context branch.

\begin{figure*}[!htbp]
    \centering
    \vspace*{-0.8cm}
    \includegraphics[width=0.73\linewidth, page=2]{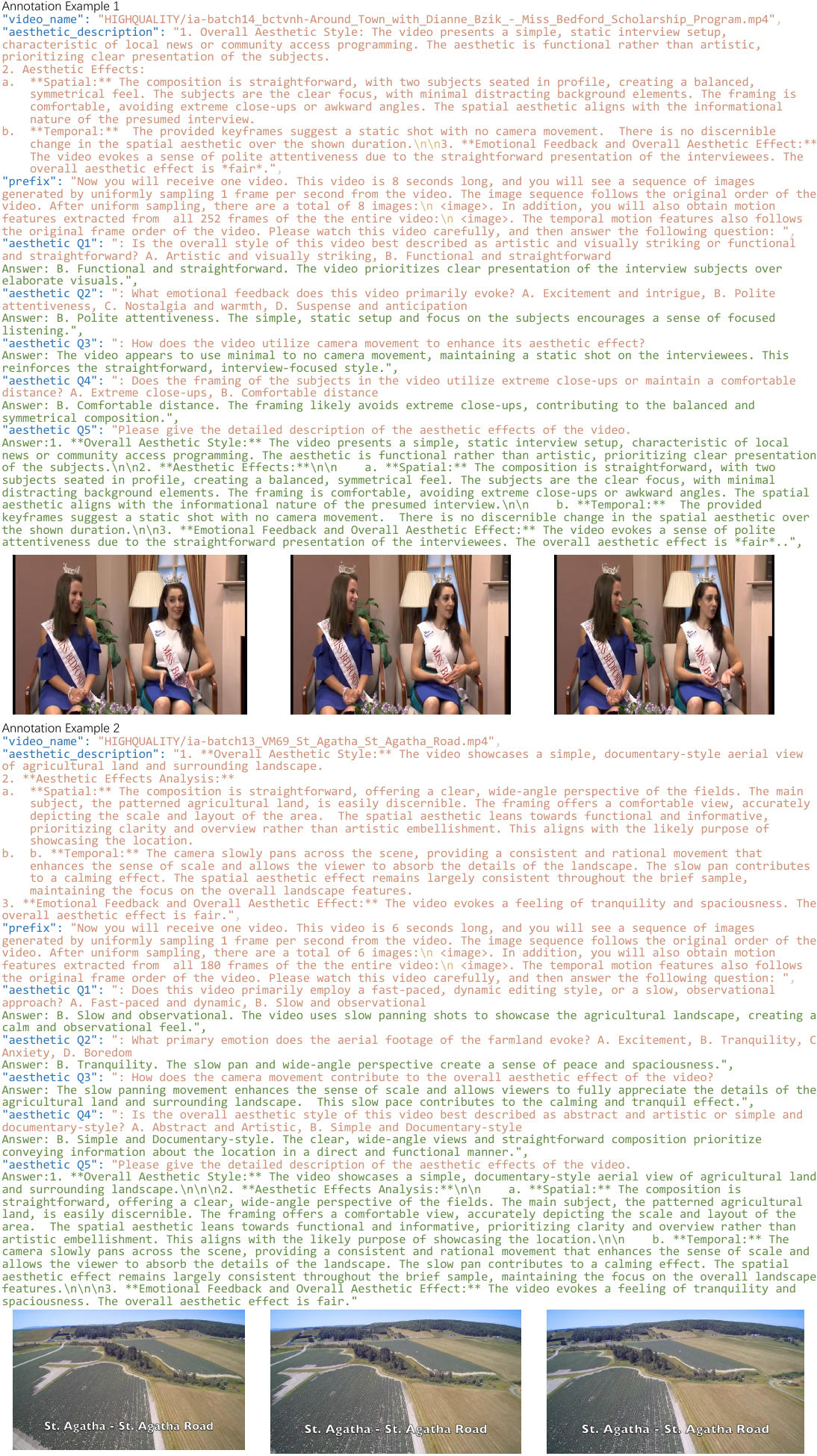}
    \caption{Aesthetic branch annotation examples.}
    \label{fig:aesthetic-branch}
\end{figure*}

\paragraph{Aesthetic branch examples}
Fig.~\ref{fig:aesthetic-branch} illustrates $2$ annotated examples from the aesthetic branch.
\section{Subjective Experiments}
\subsection{Subjective quality rating experiment}
\begin{figure*}[h]
    \centering
    \vspace*{-3cm}
    \includegraphics[width=\linewidth, page=1]{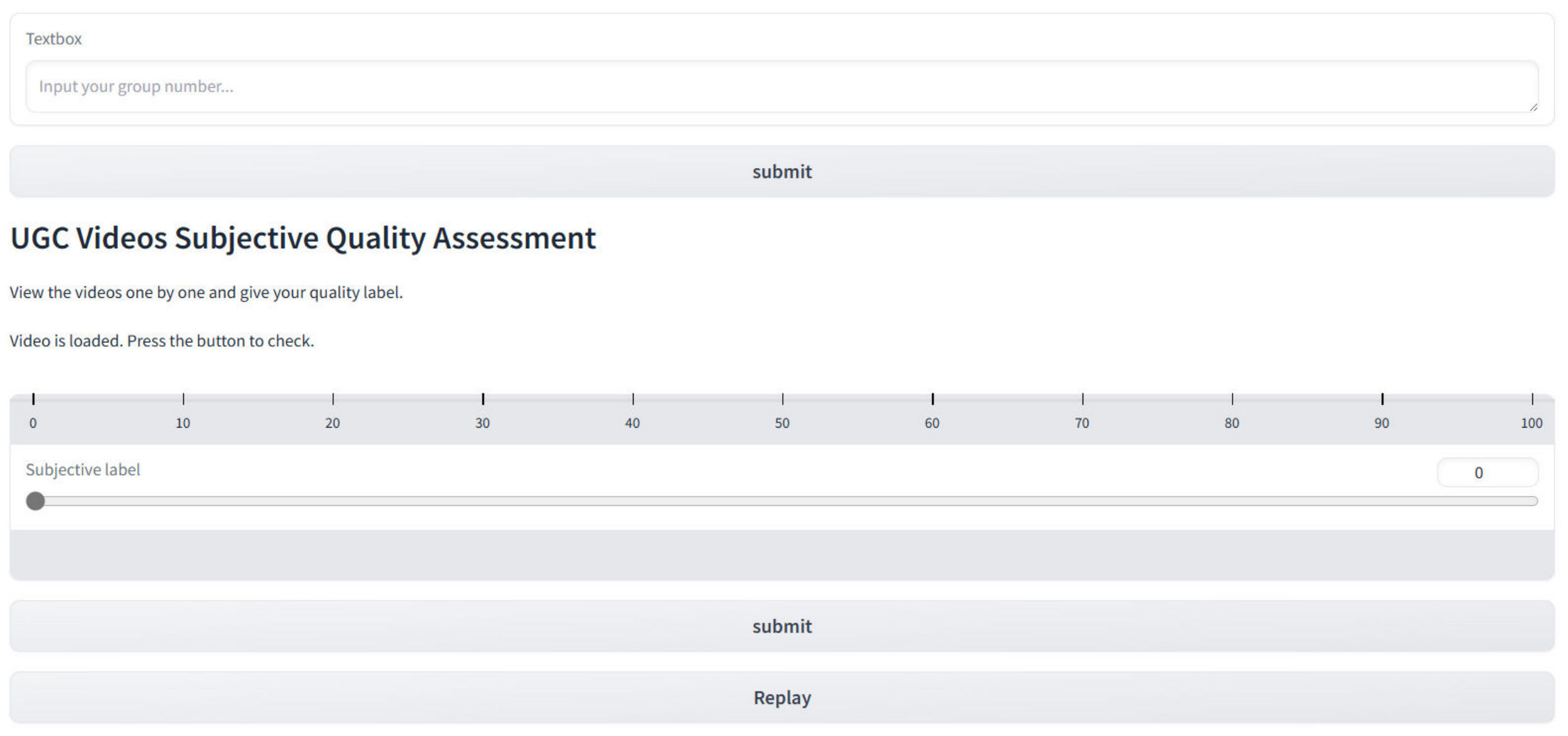}
    \caption{\textbf{Subjective Quality Rating Experiment Interface.}
    This interface is designed for the subjective quality rating experiment. The annotation workflow consists of $3$ explicit phases. Firstly, the annotator inputs their assigned group number into the designated textbox (top-center position) and activates the evaluation sequence by clicking the ``Submit" button.
    Secondly, the system plays videos from the selected group sequentially in the central viewport using a player. For each video, the annotator manipulates the horizontal slider (labeled 0 to 100; 0: "Lowest Quality", 100: ``Highest Quality") to indicate perceived quality and record their rating by clicking ``Submit". The system automatically triggers the next video in the group after each submission until all videos in the group have been rated. The interface incorporates a pre-scored range generated by a large language model. If the subjective rating falls outside this hidden range, the annotator is prompted to reconsider and rescore the video. A ``Replay" button is available for re-viewing the video.
.}
    \label{fig:Subjective quality rating experiment Interface}
\end{figure*}

To establish a rigorous framework for perceptual evaluation of UGC video quality, we develop an annotation interface illustrated in Fig. \ref{fig:Subjective quality rating experiment Interface}. The interface comprises several key elements, including a designated textbox for group number entry, a continuous rating slider for quantifying perceived quality, and a ``Replay" function for repeated video inspection. Furthermore, the system integrates a pre-determined quality range. The system will trigger a prompt for re-evaluation if a rating falls outside predefined parameters.

The subjective quality rating experiment is conducted in the standard laboratory environment, with each testing station equipped with $2$ 1080p resolution display devices. The video and rating interface are presented on separate devices. A total of $52$ participants take part in the subjective evaluation, ensuring that each video is rated by at least $10$ subjects.

\subsection{Human-in-the-loop selection experiment}
\label{Human-in-the-loop}
\begin{figure*}[htbp]
    \centering
    \vspace*{-1cm}
    \includegraphics[width=\linewidth, page=2]{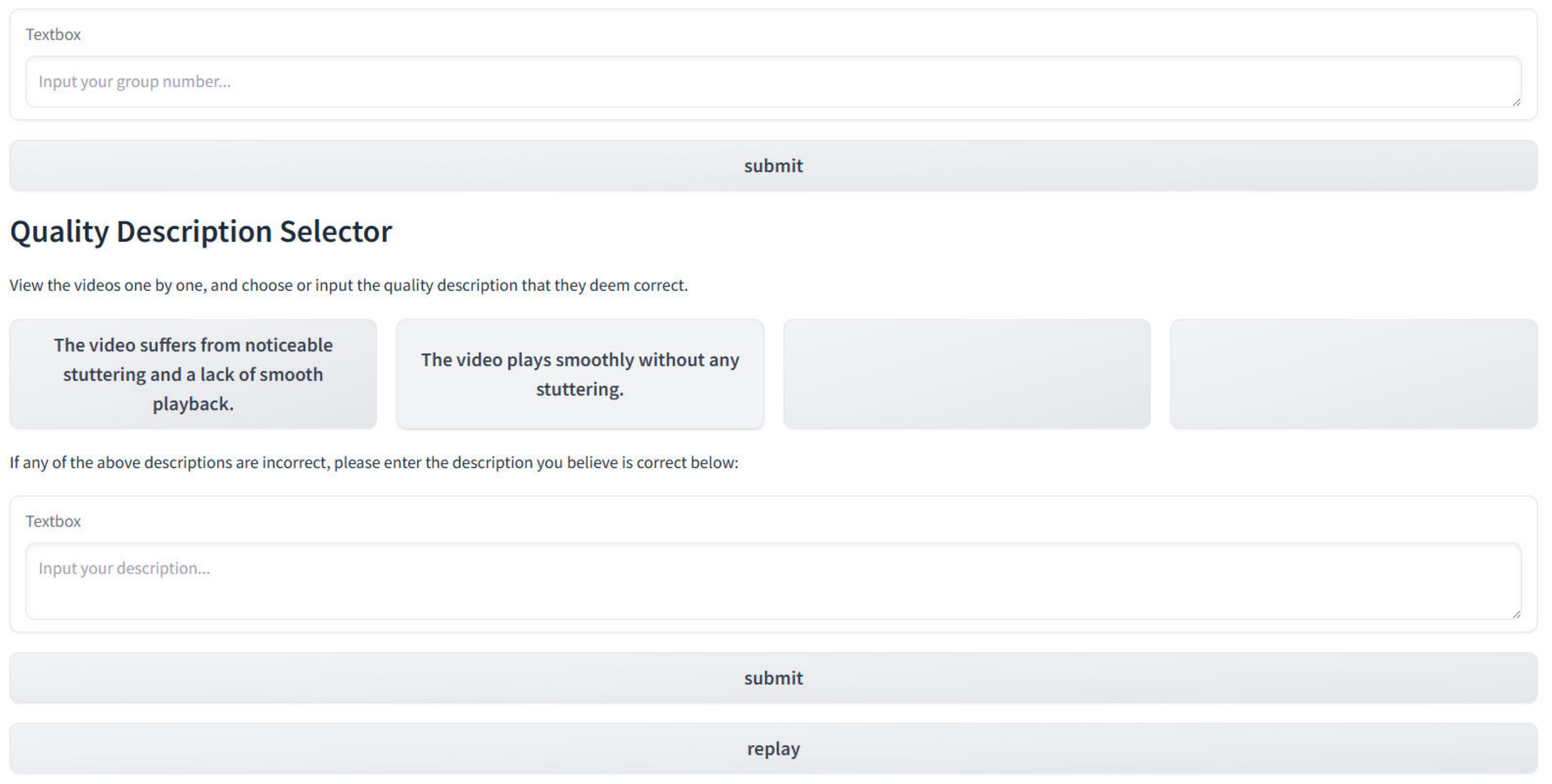}
    \caption{\textbf{Human-in-the-loop Selection Experiment Interface.}This interface facilitates a human-in-the-loop approach to video quality assessment. The annotation workflow is composed of three phases. Firstly, the annotator enters their assigned group number into the first designated textbox and activates the evaluation sequence by clicking the ``Submit" button. Secondly, the system displays videos from the selected group sequentially in the central viewport. For each video, the annotator selects the most appropriate quality description from a set of predefined options. If none of the provided options accurately reflect the perceived video quality, the annotator can manually input a custom description in the second designated textbox. Thirdly, the annotator records their selection (or custom description) by clicking the ``Submit" button. The system automatically triggers the next video in the group after each submission, continuing until all videos within the group have been assessed.
.}
    \label{fig:Human-in-the-loop selection experiment Interface}
\end{figure*}

To facilitate a human-in-the-loop methodology for video quality assessment, we implemented an interactive interface, depicted in Fig. \ref{fig:Human-in-the-loop selection experiment Interface}. The interface incorporates several essential components, including an input field for group number specification, a suite of predefined quality options, and a text field enabling manual quality assessment annotation. Additionally, the system features sequential video presentations within designated groups.

The human-in-the-loop selection experiment is conducted under the same environmental and equipment conditions as the subjective quality rating experiment. A total of $32$ expert-level participants are recruited, and the experiment is carried out after the sota LMM voting process is completed. 

In total, $11,500$ out of $23,860$ videos have at least one quality factor receiving a score of $0$ in at least one voting round, requiring manual selection. Each of these videos, along with the problematic quality factors, is carefully reviewed by at least $3$ participants. In cases where the results are inconsistent, majority voting is used to determine the final selection. 

After completing the selection process, the human-selected quality dimension results are merged back into the original annotations. Subsequently, these videos undergo quality factors summarization and the Q\&A pairs generation.

\section{Benchmark Supplementary Matrials}

\subsection{Human annotation interface}

\begin{figure*}[t]
    \centering
    \vspace*{-1cm}
    \includegraphics[width=\linewidth, page=1]{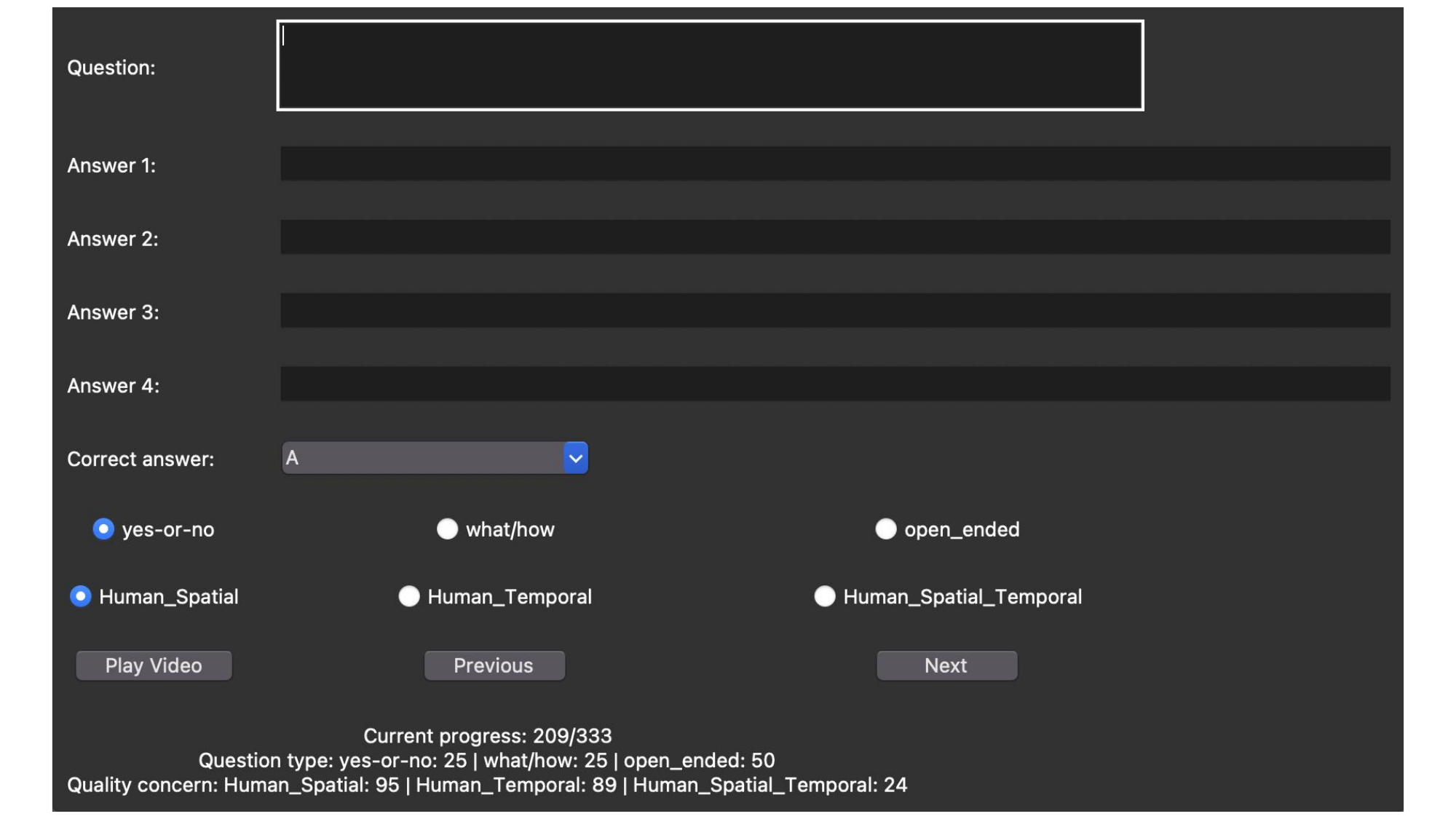}
    \caption{This is a human annotation interface with the following structure: The first row displays [Question], where the annotator fills in questions related to video distortion. [Answer 1-4] correspond to the answers for each question. For Binary type questions, only [Answer 1] and [Answer 2] need to be filled; for Multi-choice (single-answer) type questions, all four answer options should be filled in; and for Open-ended questions, only [Answer 1] is required. The [Correct Answer] is a drop-down menu where the annotator selects the correct option. Each question has corresponding question types (question type) and quality concerns (quality concern), and annotators need to choose the appropriate options based on the type of the question. Annotators can click [Play Video] to play the video, [Previous] to select the previous video, and [Next] to select the next video. The interface displays statistics at the bottom, including [Current progress], which shows the progress of the current annotated video; [Question type], which indicates the number of different types of questions; and [Quality Concern], which shows the number of different quality concerns. This helps annotators pay attention to the distribution of different annotation types during the process.}
    \label{fig:Interface}
\end{figure*}

To facilitate human annotation of low-quality videos, we design a video annotation interface, as shown in Fig. \ref{fig:Interface}. The interface consists of key components, including a section for annotators to input questions related to video distortion, corresponding answer fields based on different question types, and a selection for the correct answer. Additionally, annotators classify each question by type and quality concern. The interface also provides progress tracking and statistical summaries to help maintain a balanced distribution of annotations.
\subsection{Detailed Annotation Examples}

\begin{figure*}[!htbp]
    \centering
    \vspace*{-1cm}
    \includegraphics[width=0.7\linewidth, page=1]{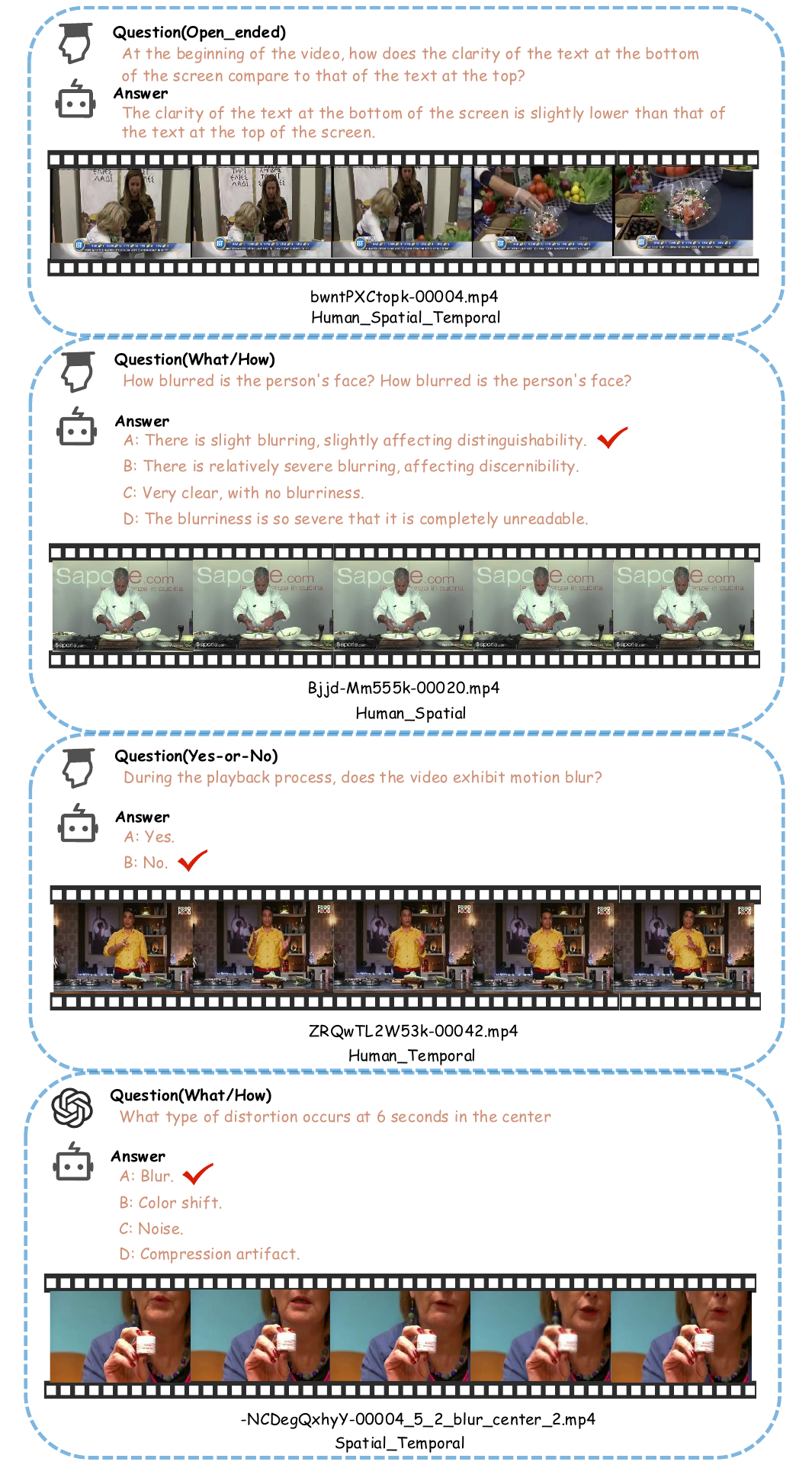}
    \caption{Examples of questions and correct answers of the OmniVQA-FG-Benchmark}
    \label{fig:benchmark(1)}
\end{figure*}

\begin{figure*}[!htbp]
    \centering
    \vspace*{-1cm}
    \includegraphics[width=0.7\linewidth, page=2]{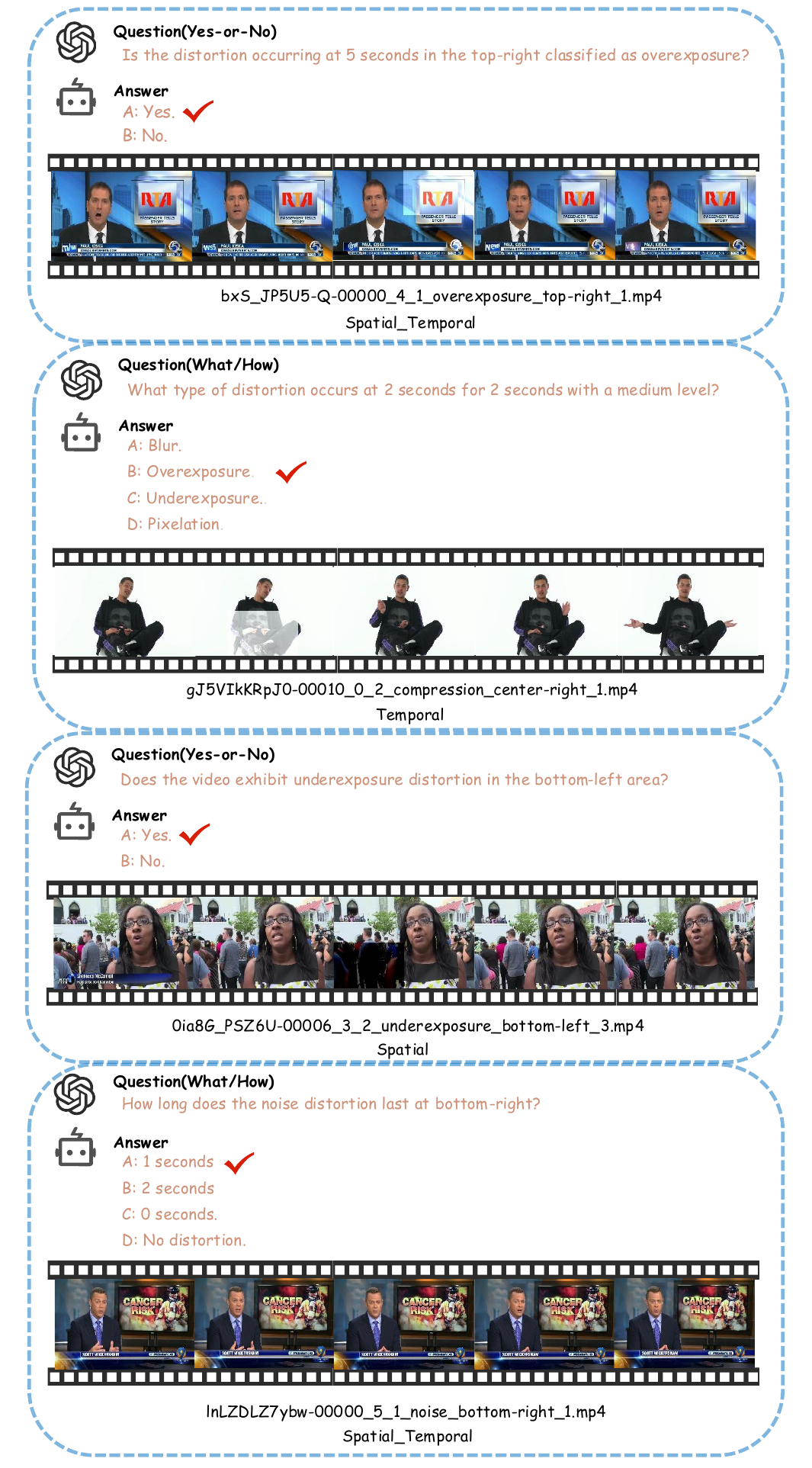}
    \caption{Examples of questions and correct answers of the OmniVQA-FG-Benchmark}
    \label{fig:benchmark(2)}
\end{figure*}

In Fig. \ref{fig:benchmark(1)} and Fig. \ref{fig:benchmark(2)}, we present specific annotation examples from the OmniVQA-FG-Benchmark, including three human-annotated cases and five machine-annotated cases. These annotations focus on different video quality concerns: spatial, temporal, and spatiotemporal aspects. Human annotations include three types of questions: Yes-or-No, What/How, and Open-ended, while machine annotations consist of two types: Yes-or-No and What/How.
\section{Applications}
\subsection{Potential future applications}
The most practical potential value of \textbf{OmniVQA} is its ability to expand the perceptual visual quality assessment MIDB at almost \textbf{zero} cost, not only in the video domain but also in the image domain. By replacing the paid \textit{sota LMM} and \textit{Gemini} with the most advanced open-source LMMs for local deployment and substituting the \textit{payed model} with a locally deployed reasoning model, the annotation costs can be nearly reduced to zero. This allows for the rapid creation of a large-scale MIDB that rivals manual annotations, with minimal sacrifice in data quality. Additionally, \textbf{OmniVQA} enables \textbf{iterative rejection sampling}, which means that after acquiring a more performant model, the original expert model in the technical branch can be replaced with a newer version to continue rejection sampling, thereby iteratively improving the data quality of the MIDB. Moreover, the human-in-the-loop process allows for data selected by human participation to be used for \textbf{RLHF} training, such as \textbf{
DPO}, to further enhance the model's ability to make accurate annotations.
\subsection{Applications in real-world scenarios}
\begin{figure*}[t]
    \centering
    \vspace*{-0.8cm}
    \includegraphics[width=\linewidth]{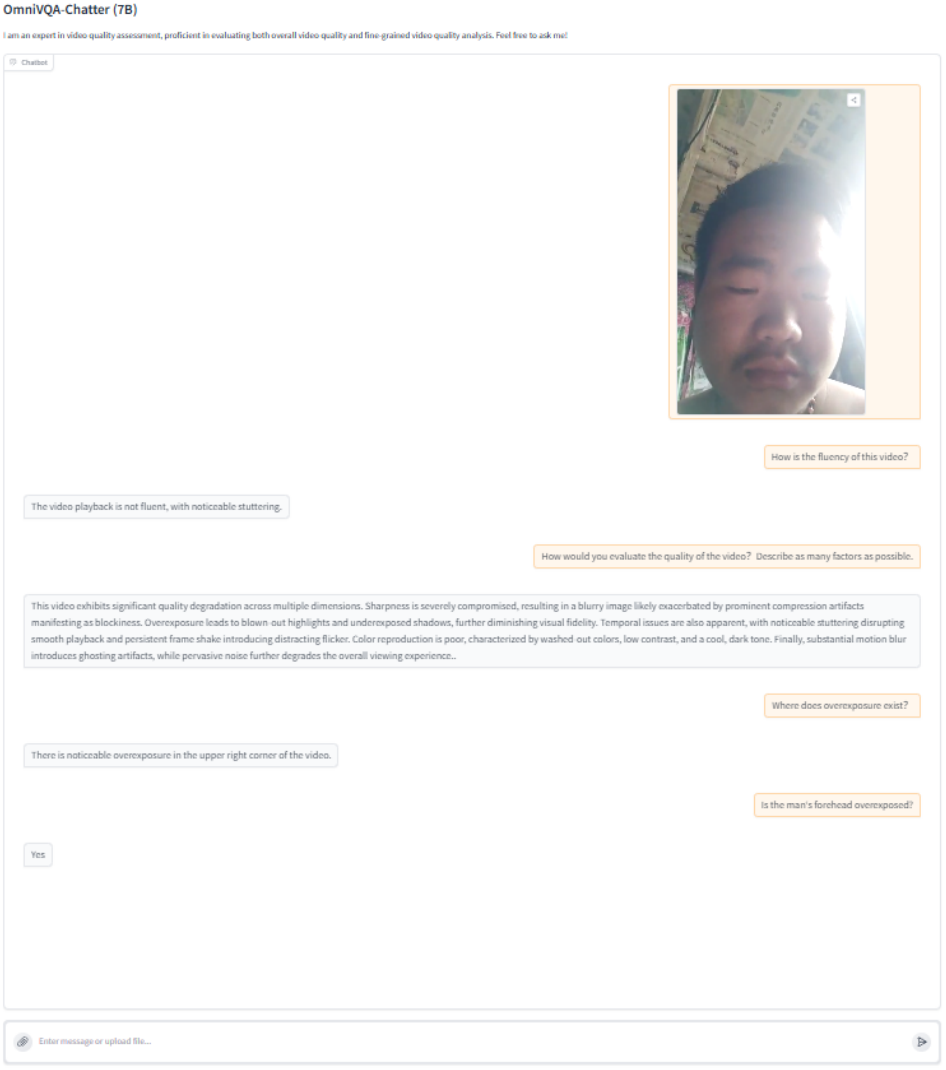}
    \caption{This is a self-recorded video excerpt from the LBVD dataset. The video exhibits severe overexposure in the upper right corner and suffers from significant stuttering, resulting in low overall quality. This example demonstrates that OmniVQA-Chatter can provide a relatively accurate and comprehensive overall video quality description while also achieving precise spatial localization of distortions.
}
    \label{fig:case1}
\end{figure*}

\begin{figure*}[hb]
    \centering
    \vspace*{-0.8cm}
    \includegraphics[width=\linewidth]{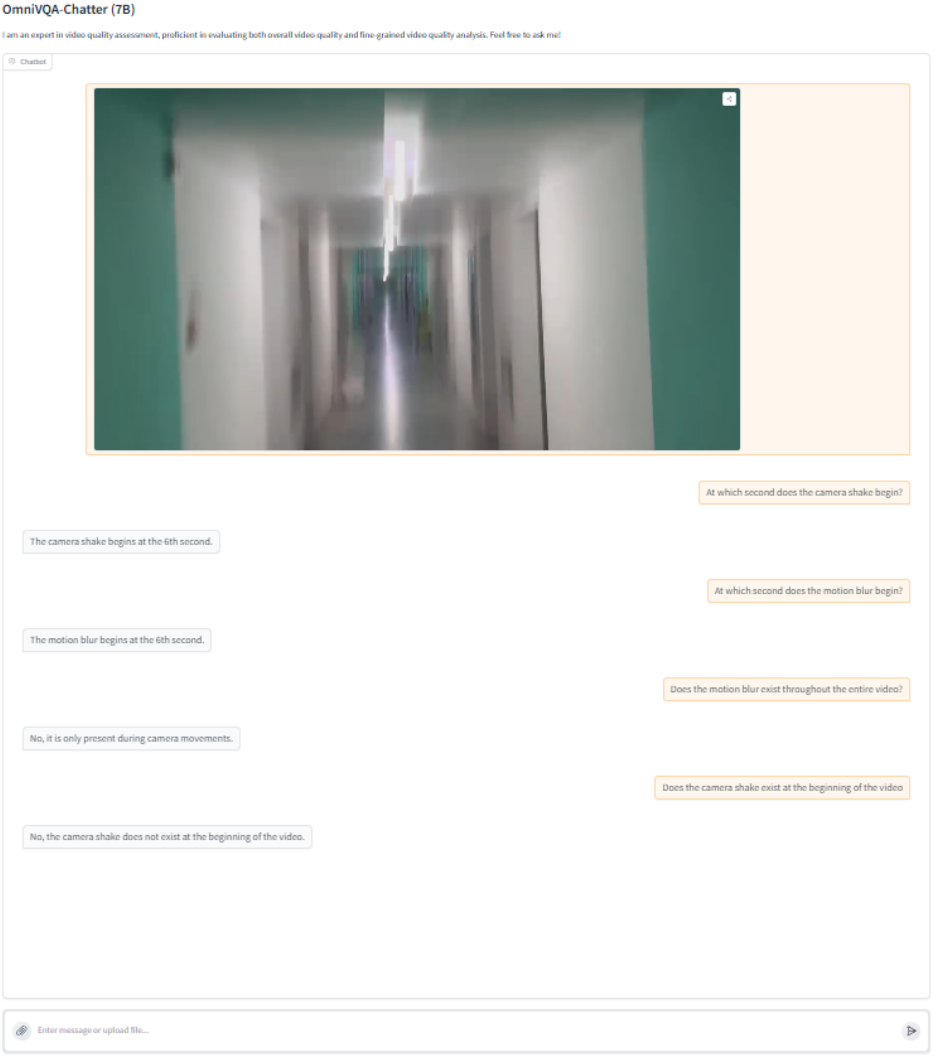}
    \caption{This is a smartphone-recorded video. For most of the playback duration, the camera remains stable; however, at the beginning of the sixth second, severe camera shake and motion blur suddenly occur. This example demonstrates that OmniVQA-Chatter possesses accurate temporal distortion localization (precise to the second)and is highly sensitive to temporal retrieval of specific distortions.
}
    \label{fig:case2}
\end{figure*}

\begin{figure*}[t]
    \centering
    \vspace*{-0.8cm}
    \includegraphics[width=\linewidth]{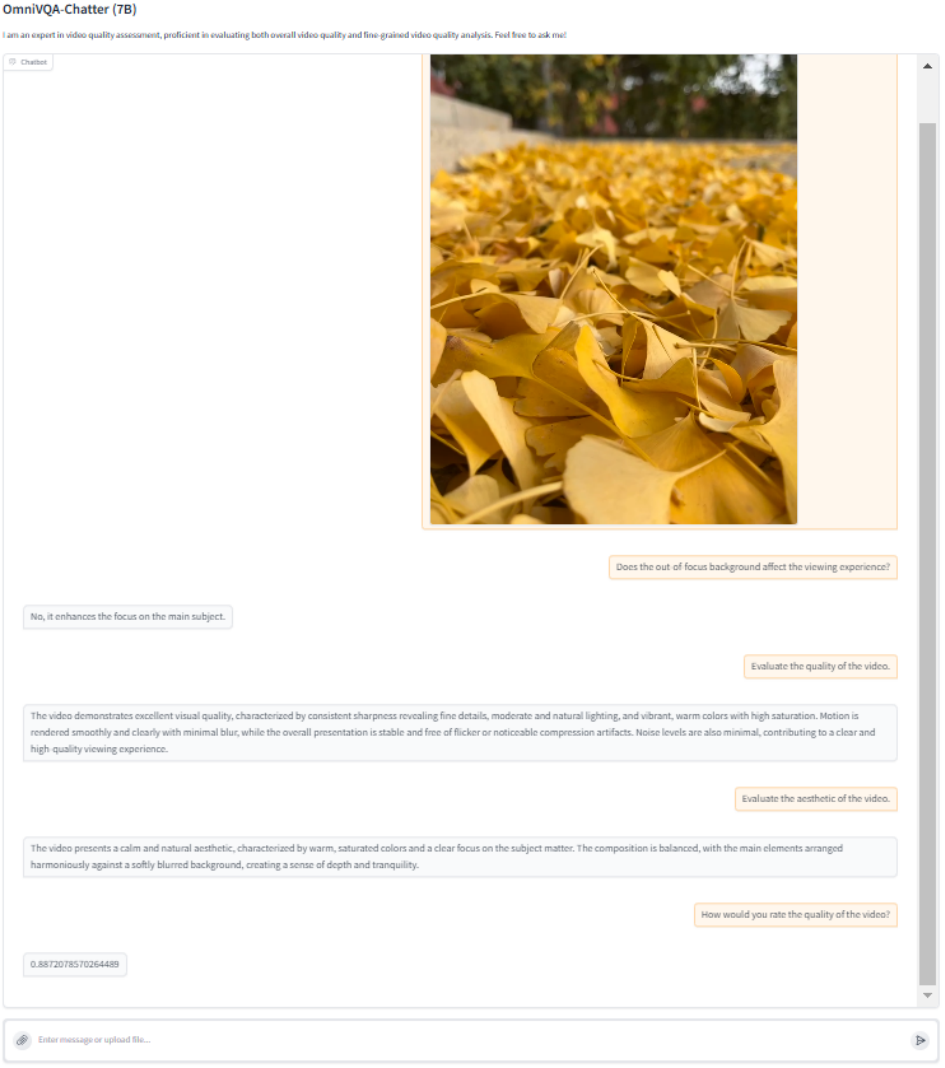}
    \caption{This video was recorded using a professional camera with background defocus processing to enhance its aesthetic appeal. The overall quality is very high, with excellent clarity. This example demonstrates that OmniVQA-Chatter can provide precise quality descriptions and quantitative scoring for high-quality videos while accurately assessing aesthetic value. Most importantly, it can analyze specific quality issues in the context of the video content (in this case, background defocus) to determine whether they should be classified as distortions. This highlights the model’s capability in ``high-level quality understanding".
}
    \label{fig:case3}
\end{figure*}

\begin{figure*}[t]
    \centering
    \vspace*{-0.8cm}
    \includegraphics[width=\linewidth]{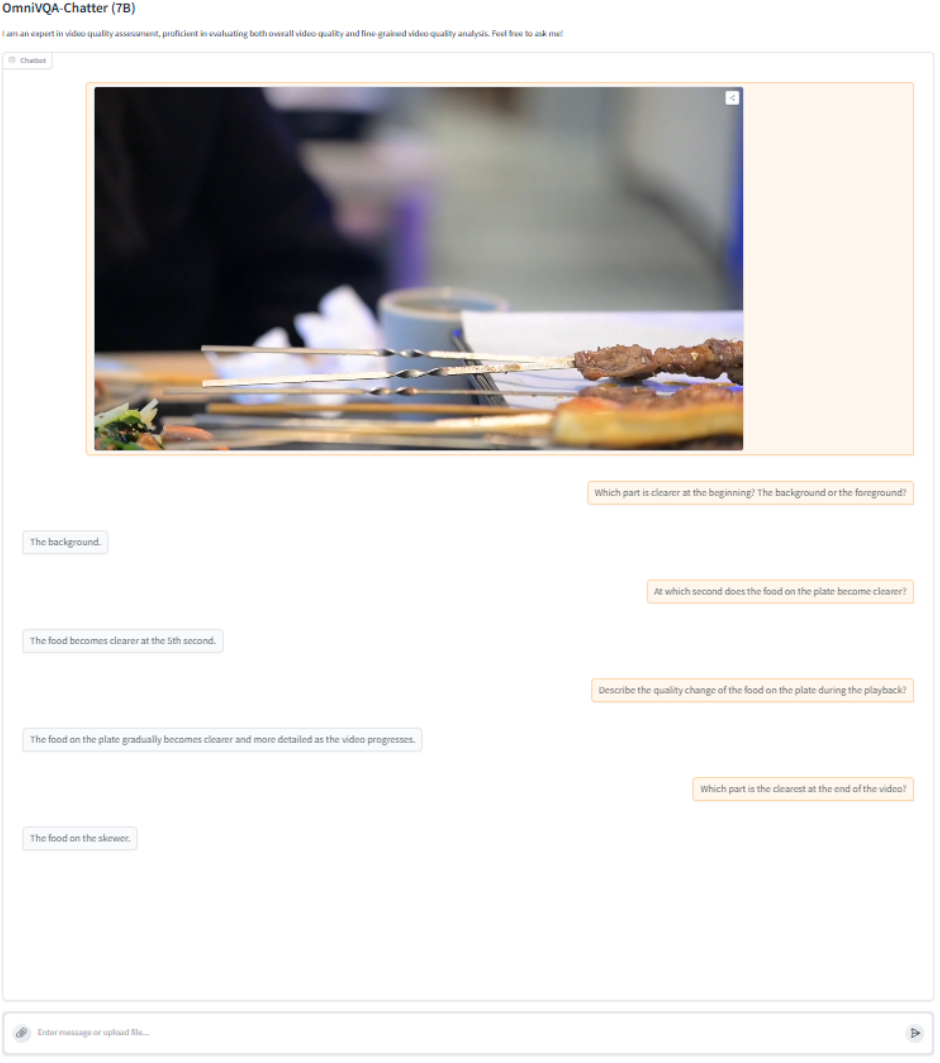}
    \caption{This video is manually recorded using a smartphone, with varying focal points over time. At the beginning of the video, the focus is primarily on the background, while the foreground appears blurred. At the fifth second, the focus shifts toward the foreground, causing the background to become blurred. This example demonstrates that OmniVQA-Chatter can perform temporal analysis of quality variations in specific regions or semantic objects, as well as precise temporal quality analysis for specific areas, accurate to the second.
}
    \label{fig:case4}
\end{figure*}

Figs. \ref{fig:case1}, \ref{fig:case2}, \ref{fig:case3}, \ref{fig:case4} show $4$ \textbf{real-world scenario} case studies presented by the \textbf{Gradio} demo.
\section{Limitations}
The primary limitation of the \textbf{OmniVQA} is that, due to computational resource constraints, we are unable to verify the \textbf{scaling effect of model parameter size in VQA under sufficiently large-scale data volume condition}. Additionally, although our approach is predominantly machine-annotation-driven, the human-in-the-loop process still incurs a certain amount of time and cost, especially in the quality rating task, where reliance on human scoring remains unavoidable in our work. These limitations hinder a comprehensive investigation into the feasibility of a \textbf{fully machine-annotated MIDB}, while effectively reducing dependence on a large amount of human annotation remains a crucial challenge in the visual quality assessment field. Addressing these limitations will be a key focus and inspiration for our future work.
\section{Acknowledgements}
We would like to express our sincere gratitude to all participants for their contributions to the human-in-the-loop subjective experiments. 





\vspace{-10pt}
\end{document}